\documentclass[11pt]{article}
\usepackage[]{acl}

\usepackage[T1]{fontenc}
\usepackage[utf8]{inputenc}
\usepackage{times}
\usepackage{latexsym}
\usepackage{microtype}
\usepackage{inconsolata}

\usepackage{graphicx}
\usepackage{amsmath,amssymb}
\usepackage{bm}
\usepackage{booktabs}
\usepackage{multirow}
\usepackage{array}
\usepackage{tabularx}
\usepackage{placeins}
\usepackage{subcaption}
\usepackage{float}
\usepackage{stfloats}
\usepackage[table]{xcolor}
\usepackage{enumitem}
\usepackage{pifont}
\usepackage{colortbl}
\usepackage{afterpage}
\usepackage[most]{tcolorbox}
\usepackage{xurl}
\usepackage{hyperref}
\usepackage{cleveref}

\newcolumntype{R}{>{\raggedleft\arraybackslash}X}
\newcolumntype{C}{>{\centering\arraybackslash}X}

\hypersetup{
  colorlinks=true,
  linkcolor=blue!50!black,
  citecolor=blue!50!black,
  urlcolor=blue!50!black
}

\definecolor{lsGreen}{HTML}{1B9E77}
\definecolor{lsOrange}{HTML}{D95F02}
\definecolor{lsGray}{HTML}{8A8A8A}

\newcommand{\LSyes}{\textcolor{lsGreen}{\ding{51}}}
\newcommand{\LSpart}{\textcolor{lsOrange}{$\blacktriangle$}}
\newcommand{\LSno}{\textcolor{lsGray}{\ding{55}}}

\setlist{topsep=2pt,partopsep=0pt,parsep=0pt,itemsep=2pt,leftmargin=*}
\setlist[itemize]{label={\textbullet},topsep=2pt,itemsep=2pt}
\setlist[enumerate]{topsep=2pt,itemsep=2pt}

\definecolor{caseEpisode}{HTML}{1F4E79}
\definecolor{caseEpisodeBg}{HTML}{E7EEF5}
\definecolor{casePre}{HTML}{2C7A7B}
\definecolor{casePreBg}{HTML}{E3F2F1}
\definecolor{casePivot}{HTML}{B7791F}
\definecolor{casePivotBg}{HTML}{FFF6E1}
\definecolor{casePost}{HTML}{6B46C1}
\definecolor{casePostBg}{HTML}{EFE8FA}
\definecolor{caseQuestion}{HTML}{2D3748}
\definecolor{caseQuestionBg}{HTML}{EDEDED}
\definecolor{caseGold}{HTML}{276749}
\definecolor{caseGoldBg}{HTML}{E2F0E5}
\definecolor{caseOptionBg}{HTML}{F7F7F8}

\graphicspath{{figures/}}

\definecolor{rowshade}{HTML}{F0F4F8}
\definecolor{groupshade}{HTML}{E5ECF4}
\definecolor{accentcell}{HTML}{D8E5F0}

\title{LongSocialBench: Do Long-Context LLMs Understand \\ Online Discussion Threads?}

\author{
Xinyi Liu \\
University of Illinois \\
Urbana-Champaign \\
\texttt{liu323@illinois.edu} \\
\And
Rinat Khaziev \\
Semantic Code Lab \\
\texttt{rinat@semantic.sh} \\
\AND
Dilek Hakkani-Tür \\
University of Illinois \\
Urbana-Champaign \\
\texttt{dilek@illinois.edu} \\
\And
Tarek F. Abdelzaher \\
University of Illinois \\
Urbana-Champaign \\
\texttt{zaher@illinois.edu} \\
}
\date{}

\begin{document}
\maketitle

\begin{abstract}
Long-context LLMs can now ingest entire online discussion threads, but understanding their social discourse requires more than reading a long document: models must track parent--reply relations, turning points, scoped subtrees, cross-branch contrasts, and participant trajectories. To test this structure-aware social reasoning, we introduce \textbf{LongSocialBench}, a benchmark of $1{,}462$ verified human-authored multiple-choice items drawn from $94$ complete Hacker News, Stack Exchange, and Reddit r/ChangeMyView episodes, with a median length of ${\approx}73$K tokens. Each item pairs a complete serialized discussion and reply structure with a four-option question, requiring models to recover structured social evidence. Released items are verified for answerability, option uniqueness, and evidence grounding.

Across $18$ models and $29$ evaluation settings, current long-context workflows remain far below human performance. The best individual result comes from Claude-Opus-4.7, which reaches $63.0\%$ when prompted to eliminate incorrect options before answering, compared with $72.4\%$ for independent human readers. Averaged across all $18$ models, the full-context Baseline scores $43.9\%$. Supplying the gold evidence scope raises this to $55.0\%$, showing that substantial errors remain even after the relevant thread region is identified. LongSocialBench shows that the missing capability is not context access or prompting alone, but social understanding over structured reply trees.
\end{abstract}

\section{Introduction}
\label{sec:intro}

\begin{figure*}[!t]
\centering
\includegraphics[width=\linewidth]{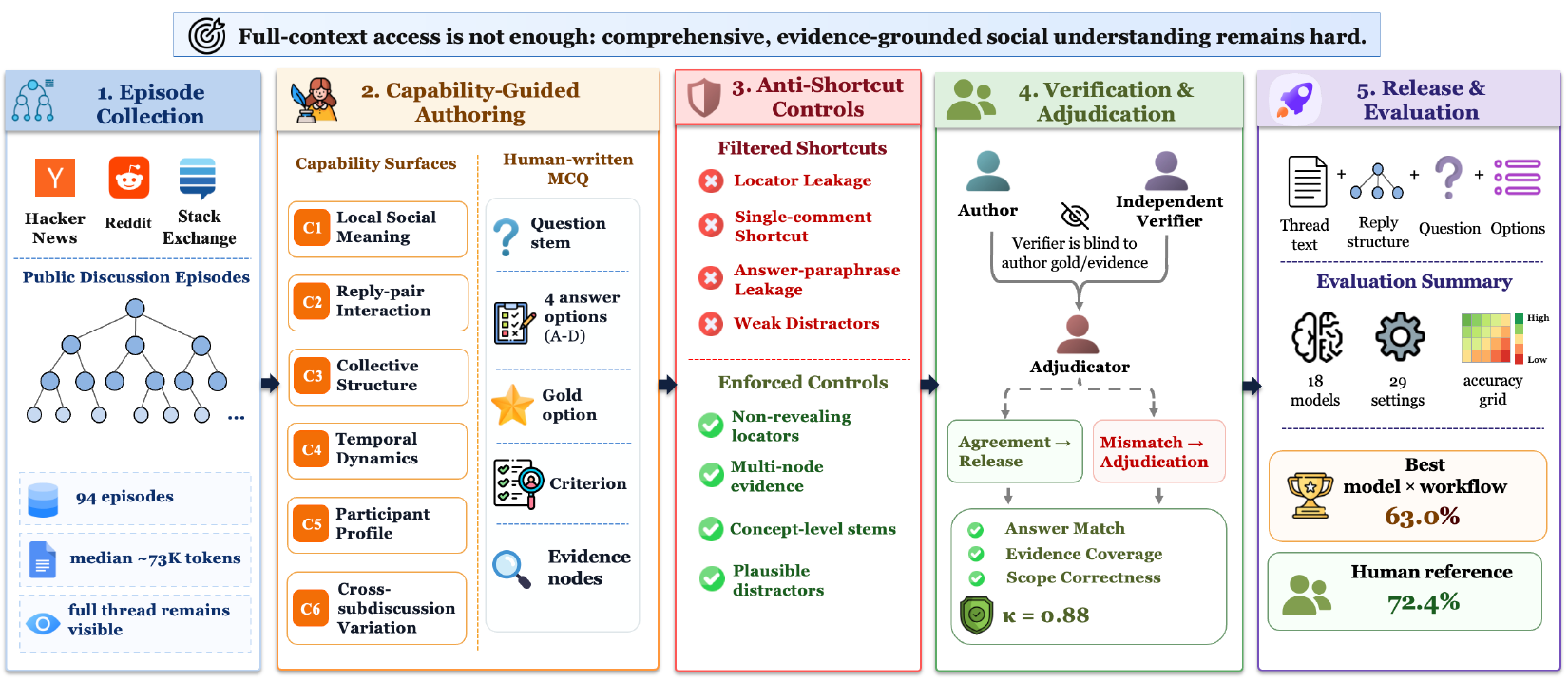}
\caption{\textbf{LongSocialBench construction and verification pipeline (\S\ref{sec:task-overview}).}
We sample long public discussions, author items guided by capability definitions, apply anti-shortcut controls, independently verify answers and evidence, and adjudicate residual disputes.}
\label{fig:pipeline-overview}
\end{figure*}

Online discussion threads are a central setting for public social interaction: participants debate claims, challenge one another, shift positions, form sub-discussions, and reveal roles through reply structure~\citep{abbott2016iac,tan2016winning,fabbri2021convosumm,niu2024mtcsd}.
Understanding the social dynamics of a discussion therefore requires more than summarizing a long document.
Relevant evidence is distributed across reply links, participant identities, branch boundaries, and temporal turning points.
A model must identify the relevant part of the thread, interpret how comments relate, and integrate evidence across the episode.
Long-context LLMs now make it possible to place a complete Hacker News thread, Stack Exchange page, or Reddit debate in context~\citep{geminiteam2024gemini15,bai2024longbench,bai2025longbenchv2,hsieh2024ruler,yen2024helmet}, but it remains unclear whether current systems can understand such discussions as structured social interaction.
This paper asks: \emph{When the complete discussion is already in context, can current LLM workflows recover the relevant evidence and determine the supported social interpretation?}

One LongSocialBench item illustrates the challenge.
In a r/ChangeMyView discussion about middle-class alienation, the item asks where the episode's unique longest reply chain first shifts from a class-and-privilege debate to a corporate wage-distribution argument.
Repeated words such as \emph{class}, \emph{privilege}, or \emph{wages} appear throughout the discussion and do not locate this transition.
The model must isolate the specified chain and identify where the exchange shifts from housing and wealth-hoarding claims to concrete claims about company revenue, GDP per worker, and worker pay.
The item therefore requires both structural localization and reasoning over how the exchange develops across turns.

Existing benchmarks cover parts of this setting, but not this combination.
Long-context benchmarks evaluate large inputs organized as documents, passages, code, tables, or synthetic structures rather than branching social interaction~\citep{shaham2023zeroscrolls,bai2024longbench,li2024loogle,bai2025longbenchv2,hsieh2024ruler,yen2024helmet}.
Social-reasoning and theory-of-mind benchmarks test pragmatic or interpersonal inference, but in short or already-localized contexts where the relevant interaction is given~\citep{sap2019socialiqa,chen2024tombench}.
Long-conversation benchmarks evaluate information retention and reasoning over extended dialogue histories, but typically within linear dialogue rather than branching, multi-party reply structures~\citep{maharana2024locomo,wu2025longmemeval}.
Online-discussion resources capture real threaded interaction, but typically formulate it as stance classification, persuasion prediction, or summarization rather than verifiable questions that require locating and reasoning over evidence in a complete reply tree~\citep{abbott2016iac,tan2016winning,fabbri2021convosumm,niu2024mtcsd}.
What remains missing is evidence-grounded social reasoning over a full reply tree, where a model must recover the relevant branch, chain, participant trajectory, or comparison set and determine which social interpretation is supported.

LongSocialBench\footnote{\url{https://huggingface.co/datasets/xinyiliu-research/LongSocialBench}}
addresses this gap with $1{,}462$ human-authored multiple-choice items drawn from $94$ complete discussions on Hacker News, Stack Exchange, and Reddit r/ChangeMyView, with a median serialized episode length of ${\approx}73$K tokens.
We use complete public discussions that fit within the evaluation context limits and span complementary forum structures and topic ranges.
Hacker News contributes technical and policy discussions, Reddit r/ChangeMyView contributes social and policy debates, and Stack Exchange contributes expert-domain Q\&A.
To define capability coverage, we draw on prior work on social inference, stance, persuasion, conversational stance, and discussion summarization~\citep{sap2019socialiqa,chen2024tombench,abbott2016iac,tan2016winning,niu2024mtcsd,fabbri2021convosumm}.
We organize recurring evidence patterns into six capability surfaces: \textbf{C1} local social meaning, \textbf{C2} reply-pair interaction, \textbf{C3} collective structure, \textbf{C4} temporal dynamics, \textbf{C5} participant profile, and \textbf{C6} cross-subdiscussion variation.
For each item, the human author specifies a decision rule for determining the correct answer from the thread and provides node-level supporting evidence.
Figure~\ref{fig:pipeline-overview} summarizes the construction and verification process.

Our evaluation reveals a substantial gap between full-context access and reliable social understanding.
We evaluate $18$ models across $29$ settings: $22$ deployable workflows that use no gold annotations, spanning full-context, reasoning, and retrieval methods, and seven diagnostic settings probing benchmark validity and evidence scope.
Across the deployable workflows, model-average accuracy ranges from $37.2\%$ to $44.7\%$, far below the $72.4\%$ mean of five independent human readers given the same full discussions.
The strongest individual model-workflow combination, Claude-Opus-4.7 prompted to eliminate incorrect options before answering, reaches $63.0\%$, still $9.4$ pp below the human mean.
As a diagnostic, we also give models the gold evidence scope to separate evidence localization from interpretation.
This raises the average across all $18$ models from $43.9\%$ under the full-context Baseline to $55.0\%$, leaving substantial error even when the relevant thread region is identified.
These results indicate difficulties in both locating relevant social evidence and interpreting it once found.

Our contributions are:
\begin{itemize}[leftmargin=*,topsep=2pt,itemsep=2pt]
\item \textbf{A human-authored benchmark for long-episode social reasoning.}
LongSocialBench contains carefully verified items written over complete real discussions and organized by six capability surfaces spanning local interaction, collective structure, participant trajectories, temporal dynamics, and cross-subdiscussion variation.

\item \textbf{A broad evaluation of long-context workflows.}
We evaluate $18$ models across $29$ settings, spanning deployable full-context, reasoning, and retrieval workflows together with validity and scope diagnostics to assess robustness across methods.

\item \textbf{A diagnosis of failure sources.}
We use gold-scope diagnostics to separate errors in locating relevant evidence from errors in interpreting it correctly, showing whether failures arise before or after the relevant evidence is identified.
\end{itemize}

Together, LongSocialBench provides a reliable and diagnostic testbed for long-context models that must understand real online discussions as structured social interaction.
\section{Related Work}
\label{sec:related}

LongSocialBench sits at the intersection of long-context evaluation and social-discourse understanding.
Prior work covers several parts of this setting, including long-input reasoning, conversational memory, localized social inference, and real online discussions, but rarely combines long contexts, reply-tree structure, social interpretation, and verifiable judgments over complete episodes.
Table~\ref{tab:prior-areas} summarizes this gap.

\begin{table}[!t]
\scriptsize
\centering
\renewcommand{\arraystretch}{1.18}
\setlength{\tabcolsep}{2.4pt}
\caption{\textbf{Area-level comparison of related evaluation settings} across long context, reply-tree structure, social interpretation, and verifiable evaluation.}
\label{tab:prior-areas}
\rowcolors{2}{rowshade}{white}
\begin{tabular*}{\linewidth}{@{\extracolsep{\fill}}lcccc@{}}
\toprule
\rowcolor{white}
\textbf{Area}            & \textbf{Long} & \textbf{Thread} & \textbf{Social} & \textbf{Verif.} \\
\midrule
Long-context eval./agg.  & \LSyes        & \LSno           & \LSno           & \LSyes \\
Long-memory dialogue     & \LSyes        & \LSno           & \LSpart         & \LSyes \\
Social commonsense/ToM   & \LSno         & \LSno           & \LSyes          & \LSyes \\
Argument / stance        & \LSpart       & \LSpart         & \LSyes          & \LSpart \\
Discussion summarization & \LSpart       & \LSpart         & \LSpart         & \LSno \\
\rowcolor{accentcell}
\textbf{Ours}            & \LSyes        & \LSyes          & \LSyes          & \LSyes \\
\bottomrule
\end{tabular*}
\vspace{1pt}

{\footnotesize
\LSyes{} = central; \LSpart{} = partial; \LSno{} = generally absent.
Long = long-context input; Thread = reply-tree structure is central to the task; Social = social interpretation; Verif. = item-level gold-answer verification.}
\end{table}

\subsection{Long-context evaluation and conversational memory.}
Long-context benchmarks test retrieval, aggregation, multi-hop reasoning, code, structured data, and synthetic context use over large inputs
\citep{shaham2023zeroscrolls,bai2024longbench,li2024loogle,
bai2025longbenchv2,hsieh2024ruler,yen2024helmet,bertsch2025oolong}.
Long-conversation benchmarks such as LoCoMo and LongMemEval instead evaluate memory and reasoning over extended dialogue histories
\citep{maharana2024locomo,wu2025longmemeval}.
Together, these benchmarks establish the difficulty of using information distributed across long contexts.
Their inputs, however, are typically documents, structured sequences, or linear dialogue histories rather than branching multi-party reply trees.
LongSocialBench adds the requirement to recover the relevant interaction structure and reason about how participants and positions develop within it.

\subsection{Social reasoning and online discussions.}
SocialIQA~\citep{sap2019socialiqa}, MuTual~\citep{cui2020mutual}, ToMi~\citep{le2019tomi}, and ToMBench~\citep{chen2024tombench} evaluate social commonsense, dialogue, or theory-of-mind reasoning, but generally use short or localized interactions.
A complementary line studies real online discussions.
IAC 2.0~\citep{abbott2016iac}, ChangeMyView persuasion studies~\citep{tan2016winning}, RumourEval~\citep{gorrell2019rumoureval}, and MT-CSD~\citep{niu2024mtcsd} annotate stance, persuasion, or related interaction outcomes, while ConvoSumm~\citep{fabbri2021convosumm} targets discussion summarization.
These resources motivate our capability design, including reply relations, collective structure, participant trajectories, and changes across sub-discussions.
LongSocialBench turns these forms of social evidence into verified questions over complete episodes, requiring models to locate the relevant thread scope and reason over the evidence within it.
\section{Task and Construction}
\label{sec:task-construction}

\subsection{Task Overview}
\label{sec:task-overview}

A LongSocialBench item is a verifiable multiple-choice question about one complete real discussion episode.
We use \emph{episode} for the full public discussion instance, such as one Hacker News thread, one Reddit r/ChangeMyView discussion, or one Stack Exchange Q\&A page.
For each item, the model receives the episode as a serialized thread with reply links, together with a question stem and four answer options.

LongSocialBench makes whole-episode social comprehension measurable through observable interaction.
Throughout, we use \emph{social} to describe how participants respond to one another, what stances and roles they take, and how these patterns persist or change across replies, participant trajectories, and sub-discussions.
Items are grounded in observable thread behavior and do not require inferences about participants' private mental states.

Drawing on prior work on social inference, stance, persuasion, conversational stance, and discussion summarization~\citep{sap2019socialiqa,chen2024tombench,abbott2016iac,tan2016winning,niu2024mtcsd,fabbri2021convosumm}, we organize recurring interaction patterns into six capability surfaces, summarized in Table~\ref{tab:category-main}.
Together, these surfaces provide systematic coverage by linking each item to a specific evidence pattern and social-reading decision.
Appendix~\ref{app:capability-table}, Table~\ref{tab:capability-design-app} provides full decision descriptions, and Appendix~\ref{app:samples} gives a worked item for each surface.

\definecolor{capblue}{HTML}{2F5F8F}
\definecolor{capbluebg}{HTML}{EAF2FB}

\begin{table}[!htbp]
\small
\centering
\renewcommand{\arraystretch}{1.12}
\setlength{\tabcolsep}{4pt}
\caption{\textbf{Overview of the six capability surfaces.}
Each surface captures a recurring evidence pattern in reply-structured discussions.}
\label{tab:category-main}
\rowcolors{2}{rowshade}{white}
\begin{tabularx}{\columnwidth}{@{}
>{\raggedright\arraybackslash}p{2.35cm}
>{\raggedright\arraybackslash}X
@{}}
\toprule
\rowcolor{capbluebg}
\textcolor{capblue}{\textbf{Capability surface}} 
& \textcolor{capblue}{\textbf{Evidence focus}} \\
\midrule

\textbf{\textcolor{capblue}{C1:} Local social meaning}
& A comment's social role or meaning in its local context. \\

\textbf{\textcolor{capblue}{C2:} Reply-pair interaction}
& The relation between a reply and the parent comment it addresses. \\

\textbf{\textcolor{capblue}{C3:} Collective structure}
& How positions or roles are distributed across a discussion region. \\

\textbf{\textcolor{capblue}{C4:} Temporal dynamics}
& How an exchange changes over time, including shifts and turning points. \\

\textbf{\textcolor{capblue}{C5:} Participant profile}
& A participant's stance or interaction pattern across multiple comments. \\

\textbf{\textcolor{capblue}{C6:} Cross-subdiscussion variation}
& How social patterns differ or remain consistent across branches or sub-discussions. \\

\bottomrule
\end{tabularx}
\end{table}

\subsection{Data Collection}
\label{sec:data-collection}

\paragraph{Episode collection and coverage.}
We collect $94$ complete public discussion episodes from Hacker News, Reddit r/ChangeMyView, and Stack Exchange.
The sources provide complementary discussion settings: Hacker News contributes many computing, AI, economy, and policy discussions, Reddit r/ChangeMyView contributes social and policy debates, and Stack Exchange contributes expert-domain Q\&A across a broad range of fields (Appendix~\ref{app:dataset-coverage}, Table~\ref{tab:topic-coverage}).
The released set contains ${\approx}6.63$M serialized tokens, with a median episode length of ${\approx}73$K tokens and a maximum of $119$K.
Per-source length and structural statistics are reported in Appendix~\ref{app:dataset-coverage}, Table~\ref{tab:episode-stats}.

\paragraph{Item design and safeguards.}
Twenty invited contributors authored items under written guidelines designed to ensure systematic capability coverage and reliable evaluation.
For each item, the author selected one of the six capability surfaces and wrote the question stem, four options, gold option, item-specific decision criterion, and supporting evidence nodes grounded in the episode.
Appendix~\ref{app:contributor-protocol} summarizes the authoring protocol, contributor training, and role separation.

To make each answer independently verifiable, a candidate item is released only if its gold option is recoverable from the episode alone, uniquely supported under the decision criterion, and grounded in cited evidence.
The evidence must also establish any structural premises needed to answer the question, such as the relevant scope, reply relation, counting unit, participant trajectory, or comparison set.
Appendix~\ref{app:release-criteria-table}, Table~\ref{tab:release-criteria-app}, details the release criteria.

To prevent shortcut solving, the guidelines control wording, option balance, and distractor design.
Stems avoid answer-revealing locators, options use parallel wording, and distractors reflect plausible errors such as selecting the wrong scope, misreading boundaries, relying on lexical overlap, or aggregating evidence incorrectly.
To reduce external answer leakage, authors also search distinctive stem and option phrases and named events on the open web, following contamination concerns in long-context benchmark construction such as LooGLE~\citep{li2024loogle}.
Items whose answers can be recovered without the supplied episode are rewritten or excluded.
Appendix~\ref{app:anti-shortcut-controls}, Table~\ref{tab:pillars-summary-app}, details these controls.

\subsection{Verification and Release}
\label{sec:verification}

Each candidate item goes through a three-phase verification process before release.
A second contributor independently answers the item, while disagreements and author-flagged items requiring additional review are examined by a third contributor.
Appendix~\ref{app:portal} documents the annotation workflow and role separation.

\textbf{Phase 1} is authoring: a contributor writes the item, decision criterion, gold option, and evidence nodes.
\textbf{Phase 2} is independent verification: a different trained contributor sees only the model-facing input, answers the question from scratch, and checks whether the authored gold option is supported.
\textbf{Phase 3} is adjudication: a third reviewer examines verifier disagreements and author-flagged items requiring additional review, re-checking the criterion, evidence, and option uniqueness.
The reviewer then confirms the item, requires revision followed by a new Phase-2 verification pass, or removes it.

Among $1{,}553$ parseable Phase-2 candidate records, $1{,}419$ match the author gold option on the first pass, yielding \textbf{$91.37\%$} first-pass agreement before adjudication.
Of the $134$ mismatches, $43$ are retained after adjudication and $91$ unresolved cases are removed.
Phase~3 also reviews $24$ author-flagged items among the $1{,}419$ first-pass matches.
The final benchmark contains $1{,}462$ items.
A stratified audit of $100$ released items finds $98$ passes and two items requiring minor wording revisions, with no residual gold-option errors (Appendix~\ref{app:release-audit}, Table~\ref{tab:release-audit}).
Additional release details, including public-data licensing, contributor protocol, and released artifacts, are documented in Appendix~\ref{app:ethics} and Appendix~\ref{app:reproducibility}.
\section{Experimental Setup}
\label{sec:setup}

\paragraph{Models.}
We evaluate $18$ contemporary LLMs spanning closed-source frontier and strong open-weight systems, including multiple versions within several model families.
Closed-source models include Claude Opus 4.7, Claude Sonnet 4.6, 4.5, and 4, GPT-5, and Gemini 2.5 Pro.
Open-weight models include Mistral Large 3, Llama 4 Scout and Maverick, Llama 3.1 and 3.2, Qwen3-Next and Qwen3-32B, GLM-5, DeepSeek-R1, Gemma 3, and OpenAI's gpt-oss models.
This selection provides broad coverage of frontier, open-weight, and reasoning-oriented systems.
Appendix~\ref{app:model-snapshots} reports the full model list and context-window limits, while Appendix~\ref{app:workflow-settings} details workflow construction and prompt instructions.

\paragraph{Workflows.}
We evaluate $29$ settings: $22$ deployable workflows that use no gold annotations and $7$ diagnostic settings.

\textbf{Reasoning prompts.}
These settings use the same complete discussion, question, and answer options as the full-context Baseline, but add an instruction for how to reason before selecting the final answer.
Standard chain-of-thought (CoT) prompting asks for free-form step-by-step reasoning.
Evidence enumeration first identifies the relevant evidence, temporal tracking reconstructs how the exchange develops across turns, and option elimination compares the choices and rules out unsupported ones.
Self-reflection asks the model to re-check its reasoning before committing to the final answer.

\textbf{Retrieval workflows.}
These methods first select or organize a smaller portion of the discussion for answering.
BM25 and E5 retrieve comments by lexical and semantic relevance, respectively.
Anchor-note RAG supplements retrieved regions with brief model-generated notes, subtree retrieval selects connected reply subtrees while preserving their structure, and LLM-based retrieval first asks a model to select relevant comments or regions.

\textbf{Diagnostics.}
The seven diagnostic settings alter the information available to the model to test potential shortcuts, answer-format effects, and evidence localization.
Five validity probes isolate specific cues: options-only provides only the four answer choices, question-only provides only the question stem, title-and-root-only provides the discussion title and root post without replies, shuffled-thread keeps all comments but shuffles their order, and open-answer keeps the thread and question but removes the answer choices and requires a short free-text response.
Two scope diagnostics test whether errors come from failing to find the relevant evidence: gold-scope provides the author-annotated evidence scope, while infer-scope provides a scope predicted by a separate prompt.

Table~\ref{tab:workflow-settings} in Appendix~\ref{app:workflow-settings} gives the exact settings, input construction, and prompt instructions.

\paragraph{Metrics.}
For multiple-choice settings, responses are deterministically parsed to an A--D choice and scored by exact match.
Accuracy is computed over successfully parsed responses only, excluding outputs for which no unique answer choice can be recovered.
Benchmark-level results report micro accuracy, while capability summaries report the unweighted macro mean over C1--C6.
For the open-answer validity probe, free-text responses are content-scored against the gold option.
\section{Results}
\label{sec:results}

We first establish the overall performance gap, then examine how it varies across capabilities and workflows, and finally use diagnostic analyses to identify where the remaining errors arise.

\subsection{Current models remain well below human performance}
\label{sec:results-headline}
\begin{table*}[!t]
\centering
\caption{\textbf{Per-model capability accuracy under Baseline and CoT.}
Cells report accuracy; \textbf{Avg} is the C1--C6 macro mean.
CoT superscripts show lift over Baseline, with green/red for positive/negative change.
\textbf{Bold} marks the best model per column; C1--C6 are defined in Table~\ref{tab:capability-design-app}.}
\label{tab:capability-main}
\resizebox{\textwidth}{!}{%
\renewcommand{\arraystretch}{1.10}
\setlength{\tabcolsep}{4pt}
\begin{tabular}{@{}l rrrrrr | r | rrrrrr | r@{}}
\toprule
 & \multicolumn{7}{c|}{\textbf{Baseline (Full-context)}} & \multicolumn{7}{c}{\textbf{CoT (with $\Delta$ vs Baseline)}} \\
\cmidrule(lr){2-8}\cmidrule(lr){9-15}
\textbf{Model}     & C1            & C2            & C3            & C4            & C5            & C6            & Avg           & C1                                                                             & C2                                                                              & C3                                                                             & C4                                                                             & C5                                                                              & C6                                                                             & Avg \\
\midrule
\multicolumn{15}{@{}l}{\emph{(Closed-source frontier)}} \\
Claude-Opus-4.7    & \textbf{74.5} & 65.2          & 49.1          & \textbf{67.0} & 54.1          & 50.0          & \textbf{60.0} & \textbf{79.4}\,\raisebox{0.6ex}{\scriptsize\textcolor{green!55!black}{$+4.8$}} & 71.9\,\raisebox{0.6ex}{\scriptsize\textcolor{green!55!black}{$+6.7$}}           & 53.1\,\raisebox{0.6ex}{\scriptsize\textcolor{green!55!black}{$+4.1$}}          & \textbf{70.8}\,\raisebox{0.6ex}{\scriptsize\textcolor{green!55!black}{$+3.8$}} & \textbf{65.6}\,\raisebox{0.6ex}{\scriptsize\textcolor{green!55!black}{$+11.5$}} & \textbf{59.3}\,\raisebox{0.6ex}{\scriptsize\textcolor{green!55!black}{$+9.3$}} & \textbf{66.7}\,\raisebox{0.6ex}{\scriptsize\textcolor{green!55!black}{$+6.7$}} \\
Gemini-2.5-Pro     & 65.5          & 58.4          & \textbf{52.1} & 49.7          & 45.9          & 36.4          & 51.3          & 71.5\,\raisebox{0.6ex}{\scriptsize\textcolor{green!55!black}{$+6.1$}}          & 69.7\,\raisebox{0.6ex}{\scriptsize\textcolor{green!55!black}{$+11.2$}}          & \textbf{55.9}\,\raisebox{0.6ex}{\scriptsize\textcolor{green!55!black}{$+3.8$}} & 56.8\,\raisebox{0.6ex}{\scriptsize\textcolor{green!55!black}{$+7.0$}}          & 56.9\,\raisebox{0.6ex}{\scriptsize\textcolor{green!55!black}{$+11.0$}}          & 56.8\,\raisebox{0.6ex}{\scriptsize\textcolor{green!55!black}{$+20.3$}}         & 61.2\,\raisebox{0.6ex}{\scriptsize\textcolor{green!55!black}{$+9.9$}} \\
GPT-5              & 61.8          & 68.5          & 43.7          & 58.9          & \textbf{63.8} & \textbf{53.4} & 58.3          & 67.9\,\raisebox{0.6ex}{\scriptsize\textcolor{green!55!black}{$+6.1$}}          & \textbf{80.9}\,\raisebox{0.6ex}{\scriptsize\textcolor{green!55!black}{$+12.4$}} & 44.3\,\raisebox{0.6ex}{\scriptsize\textcolor{green!55!black}{$+0.6$}}          & 49.7\,\raisebox{0.6ex}{\scriptsize\textcolor{red}{$-9.2$}}                     & 63.3\,\raisebox{0.6ex}{\scriptsize\textcolor{red}{$-0.5$}}                      & 58.5\,\raisebox{0.6ex}{\scriptsize\textcolor{green!55!black}{$+5.1$}}          & 60.8\,\raisebox{0.6ex}{\scriptsize\textcolor{green!55!black}{$+2.4$}} \\
Claude-Sonnet-4.5  & 73.3          & 59.6          & 46.6          & 49.7          & 50.5          & 46.6          & 54.4          & 71.5\,\raisebox{0.6ex}{\scriptsize\textcolor{red}{$-1.8$}}                     & 67.4\,\raisebox{0.6ex}{\scriptsize\textcolor{green!55!black}{$+7.9$}}           & 46.1\,\raisebox{0.6ex}{\scriptsize\textcolor{red}{$-0.4$}}                     & 53.0\,\raisebox{0.6ex}{\scriptsize\textcolor{green!55!black}{$+3.2$}}          & 56.0\,\raisebox{0.6ex}{\scriptsize\textcolor{green!55!black}{$+5.5$}}           & 48.3\,\raisebox{0.6ex}{\scriptsize\textcolor{green!55!black}{$+1.7$}}          & 57.1\,\raisebox{0.6ex}{\scriptsize\textcolor{green!55!black}{$+2.7$}} \\
Claude-4           & 72.7          & \textbf{71.9} & 45.0          & 54.6          & 50.5          & 47.5          & 57.0          & 72.1\,\raisebox{0.6ex}{\scriptsize\textcolor{red}{$-0.6$}}                     & 71.9\,\raisebox{0.6ex}{\scriptsize\textcolor{green!55!black}{$+0.0$}}           & 46.9\,\raisebox{0.6ex}{\scriptsize\textcolor{green!55!black}{$+1.9$}}          & 51.9\,\raisebox{0.6ex}{\scriptsize\textcolor{red}{$-2.7$}}                     & 50.5\,\raisebox{0.6ex}{\scriptsize\textcolor{green!55!black}{$+0.0$}}           & 48.3\,\raisebox{0.6ex}{\scriptsize\textcolor{green!55!black}{$+0.8$}}          & 56.9\,\raisebox{0.6ex}{\scriptsize\textcolor{red}{$-0.1$}} \\
Claude-Sonnet-4.6  & 56.4          & 49.4          & 41.5          & 43.2          & 47.2          & 36.4          & 45.7          & 65.5\,\raisebox{0.6ex}{\scriptsize\textcolor{green!55!black}{$+9.1$}}          & 58.4\,\raisebox{0.6ex}{\scriptsize\textcolor{green!55!black}{$+9.0$}}           & 47.6\,\raisebox{0.6ex}{\scriptsize\textcolor{green!55!black}{$+6.1$}}          & 51.4\,\raisebox{0.6ex}{\scriptsize\textcolor{green!55!black}{$+8.1$}}          & 53.7\,\raisebox{0.6ex}{\scriptsize\textcolor{green!55!black}{$+6.4$}}           & 47.5\,\raisebox{0.6ex}{\scriptsize\textcolor{green!55!black}{$+11.0$}}         & 54.0\,\raisebox{0.6ex}{\scriptsize\textcolor{green!55!black}{$+8.3$}} \\
\midrule
\multicolumn{15}{@{}l}{\emph{(Strong open-weight)}} \\
GLM-5              & 54.5          & 36.0          & 41.0          & 35.1          & 41.3          & 39.8          & 41.3          & 72.1\,\raisebox{0.6ex}{\scriptsize\textcolor{green!55!black}{$+17.6$}}         & 64.0\,\raisebox{0.6ex}{\scriptsize\textcolor{green!55!black}{$+28.1$}}          & 47.0\,\raisebox{0.6ex}{\scriptsize\textcolor{green!55!black}{$+6.0$}}          & 52.4\,\raisebox{0.6ex}{\scriptsize\textcolor{green!55!black}{$+17.3$}}         & 59.2\,\raisebox{0.6ex}{\scriptsize\textcolor{green!55!black}{$+17.9$}}          & 45.8\,\raisebox{0.6ex}{\scriptsize\textcolor{green!55!black}{$+5.9$}}          & 56.8\,\raisebox{0.6ex}{\scriptsize\textcolor{green!55!black}{$+15.5$}} \\
DeepSeek-R1        & 64.8          & 60.7          & 42.5          & 51.4          & 45.0          & 39.0          & 50.6          & 65.5\,\raisebox{0.6ex}{\scriptsize\textcolor{green!55!black}{$+0.6$}}          & 61.8\,\raisebox{0.6ex}{\scriptsize\textcolor{green!55!black}{$+1.1$}}           & 42.4\,\raisebox{0.6ex}{\scriptsize\textcolor{red}{$-0.1$}}                     & 46.5\,\raisebox{0.6ex}{\scriptsize\textcolor{red}{$-4.9$}}                     & 46.8\,\raisebox{0.6ex}{\scriptsize\textcolor{green!55!black}{$+1.8$}}           & 39.8\,\raisebox{0.6ex}{\scriptsize\textcolor{green!55!black}{$+0.8$}}          & 50.5\,\raisebox{0.6ex}{\scriptsize\textcolor{red}{$-0.1$}} \\
Qwen3-Next-80B     & 56.4          & 49.4          & 40.2          & 44.9          & 38.5          & 38.1          & 44.6          & 64.2\,\raisebox{0.6ex}{\scriptsize\textcolor{green!55!black}{$+7.9$}}          & 52.8\,\raisebox{0.6ex}{\scriptsize\textcolor{green!55!black}{$+3.4$}}           & 37.4\,\raisebox{0.6ex}{\scriptsize\textcolor{red}{$-2.8$}}                     & 42.7\,\raisebox{0.6ex}{\scriptsize\textcolor{red}{$-2.2$}}                     & 42.7\,\raisebox{0.6ex}{\scriptsize\textcolor{green!55!black}{$+4.1$}}           & 37.3\,\raisebox{0.6ex}{\scriptsize\textcolor{red}{$-0.8$}}                     & 46.2\,\raisebox{0.6ex}{\scriptsize\textcolor{green!55!black}{$+1.6$}} \\
Mistral-Large-3    & 55.8          & 43.8          & 40.2          & 42.7          & 36.7          & 40.7          & 43.3          & 64.8\,\raisebox{0.6ex}{\scriptsize\textcolor{green!55!black}{$+9.1$}}          & 55.1\,\raisebox{0.6ex}{\scriptsize\textcolor{green!55!black}{$+11.2$}}          & 37.8\,\raisebox{0.6ex}{\scriptsize\textcolor{red}{$-2.3$}}                     & 34.6\,\raisebox{0.6ex}{\scriptsize\textcolor{red}{$-8.1$}}                     & 45.9\,\raisebox{0.6ex}{\scriptsize\textcolor{green!55!black}{$+9.2$}}           & 38.1\,\raisebox{0.6ex}{\scriptsize\textcolor{red}{$-2.5$}}                     & 46.1\,\raisebox{0.6ex}{\scriptsize\textcolor{green!55!black}{$+2.8$}} \\
\midrule
\multicolumn{15}{@{}l}{\emph{(Mid open-weight)}} \\
Llama-4-Mav-17B    & 55.8          & 42.7          & 40.9          & 37.8          & 37.6          & 33.9          & 41.5          & 58.8\,\raisebox{0.6ex}{\scriptsize\textcolor{green!55!black}{$+3.0$}}          & 62.9\,\raisebox{0.6ex}{\scriptsize\textcolor{green!55!black}{$+20.2$}}          & 41.0\,\raisebox{0.6ex}{\scriptsize\textcolor{green!55!black}{$+0.1$}}          & 38.9\,\raisebox{0.6ex}{\scriptsize\textcolor{green!55!black}{$+1.1$}}          & 44.5\,\raisebox{0.6ex}{\scriptsize\textcolor{green!55!black}{$+6.9$}}           & 30.5\,\raisebox{0.6ex}{\scriptsize\textcolor{red}{$-3.4$}}                     & 46.1\,\raisebox{0.6ex}{\scriptsize\textcolor{green!55!black}{$+4.7$}} \\
Llama-4-Scout-17B  & 55.2          & 47.2          & 38.7          & 33.5          & 39.9          & 33.1          & 41.3          & 53.9\,\raisebox{0.6ex}{\scriptsize\textcolor{red}{$-1.2$}}                     & 49.4\,\raisebox{0.6ex}{\scriptsize\textcolor{green!55!black}{$+2.2$}}           & 39.7\,\raisebox{0.6ex}{\scriptsize\textcolor{green!55!black}{$+1.0$}}          & 30.3\,\raisebox{0.6ex}{\scriptsize\textcolor{red}{$-3.2$}}                     & 39.0\,\raisebox{0.6ex}{\scriptsize\textcolor{red}{$-0.9$}}                      & 32.2\,\raisebox{0.6ex}{\scriptsize\textcolor{red}{$-0.8$}}                     & 40.8\,\raisebox{0.6ex}{\scriptsize\textcolor{red}{$-0.5$}} \\
Llama-3.1-70B      & 53.9          & 41.6          & 36.1          & 35.1          & 33.9          & 36.4          & 39.5          & 54.5\,\raisebox{0.6ex}{\scriptsize\textcolor{green!55!black}{$+0.6$}}          & 52.8\,\raisebox{0.6ex}{\scriptsize\textcolor{green!55!black}{$+11.2$}}          & 33.9\,\raisebox{0.6ex}{\scriptsize\textcolor{red}{$-2.2$}}                     & 36.2\,\raisebox{0.6ex}{\scriptsize\textcolor{green!55!black}{$+1.1$}}          & 39.9\,\raisebox{0.6ex}{\scriptsize\textcolor{green!55!black}{$+6.0$}}           & 40.7\,\raisebox{0.6ex}{\scriptsize\textcolor{green!55!black}{$+4.2$}}          & 43.0\,\raisebox{0.6ex}{\scriptsize\textcolor{green!55!black}{$+3.5$}} \\
Qwen3-32B          & 60.6          & 51.7          & 38.9          & 37.8          & 41.7          & 32.2          & 43.8          & 51.5\,\raisebox{0.6ex}{\scriptsize\textcolor{red}{$-9.1$}}                     & 39.3\,\raisebox{0.6ex}{\scriptsize\textcolor{red}{$-12.4$}}                     & 26.8\,\raisebox{0.6ex}{\scriptsize\textcolor{red}{$-12.1$}}                    & 31.9\,\raisebox{0.6ex}{\scriptsize\textcolor{red}{$-5.9$}}                     & 28.4\,\raisebox{0.6ex}{\scriptsize\textcolor{red}{$-13.3$}}                     & 29.7\,\raisebox{0.6ex}{\scriptsize\textcolor{red}{$-2.5$}}                     & 34.6\,\raisebox{0.6ex}{\scriptsize\textcolor{red}{$-9.2$}} \\
\midrule
\multicolumn{15}{@{}l}{\emph{(Small / mid-weak)}} \\
GPT-OSS-120B       & 58.8          & 48.3          & 36.2          & 34.1          & 36.7          & 38.1          & 42.0          & 54.5\,\raisebox{0.6ex}{\scriptsize\textcolor{red}{$-4.2$}}                     & 49.4\,\raisebox{0.6ex}{\scriptsize\textcolor{green!55!black}{$+1.1$}}           & 33.2\,\raisebox{0.6ex}{\scriptsize\textcolor{red}{$-3.1$}}                     & 33.0\,\raisebox{0.6ex}{\scriptsize\textcolor{red}{$-1.1$}}                     & 37.2\,\raisebox{0.6ex}{\scriptsize\textcolor{green!55!black}{$+0.5$}}           & 32.2\,\raisebox{0.6ex}{\scriptsize\textcolor{red}{$-5.9$}}                     & 39.9\,\raisebox{0.6ex}{\scriptsize\textcolor{red}{$-2.1$}} \\
GPT-OSS-20B        & 49.7          & 46.1          & 31.6          & 36.8          & 38.1          & 33.1          & 39.2          & 49.1\,\raisebox{0.6ex}{\scriptsize\textcolor{red}{$-0.6$}}                     & 48.3\,\raisebox{0.6ex}{\scriptsize\textcolor{green!55!black}{$+2.2$}}           & 31.7\,\raisebox{0.6ex}{\scriptsize\textcolor{green!55!black}{$+0.1$}}          & 34.1\,\raisebox{0.6ex}{\scriptsize\textcolor{red}{$-2.7$}}                     & 37.6\,\raisebox{0.6ex}{\scriptsize\textcolor{red}{$-0.5$}}                      & 33.1\,\raisebox{0.6ex}{\scriptsize\textcolor{green!55!black}{$+0.0$}}          & 39.0\,\raisebox{0.6ex}{\scriptsize\textcolor{red}{$-0.2$}} \\
Llama-3.2-90B      & 50.3          & 32.6          & 35.8          & 28.6          & 33.9          & 30.5          & 35.3          & 46.7\,\raisebox{0.6ex}{\scriptsize\textcolor{red}{$-3.6$}}                     & 32.6\,\raisebox{0.6ex}{\scriptsize\textcolor{green!55!black}{$+0.0$}}           & 31.3\,\raisebox{0.6ex}{\scriptsize\textcolor{red}{$-4.5$}}                     & 25.4\,\raisebox{0.6ex}{\scriptsize\textcolor{red}{$-3.2$}}                     & 42.2\,\raisebox{0.6ex}{\scriptsize\textcolor{green!55!black}{$+8.3$}}           & 31.4\,\raisebox{0.6ex}{\scriptsize\textcolor{green!55!black}{$+0.8$}}          & 34.9\,\raisebox{0.6ex}{\scriptsize\textcolor{red}{$-0.4$}} \\
Gemma-3-27B        & 57.0          & 40.4          & 33.9          & 33.5          & 33.9          & 29.7          & 38.1          & 47.3\,\raisebox{0.6ex}{\scriptsize\textcolor{red}{$-9.7$}}                     & 44.9\,\raisebox{0.6ex}{\scriptsize\textcolor{green!55!black}{$+4.5$}}           & 25.9\,\raisebox{0.6ex}{\scriptsize\textcolor{red}{$-8.0$}}                     & 31.4\,\raisebox{0.6ex}{\scriptsize\textcolor{red}{$-2.2$}}                     & 35.3\,\raisebox{0.6ex}{\scriptsize\textcolor{green!55!black}{$+1.4$}}           & 31.4\,\raisebox{0.6ex}{\scriptsize\textcolor{green!55!black}{$+1.7$}}          & 36.0\,\raisebox{0.6ex}{\scriptsize\textcolor{red}{$-2.0$}} \\
\midrule
\rowcolor{accentcell}
\textbf{Mean (18)} & 59.8          & 50.7          & 40.8          & 43.0          & 42.7          & 38.6          & 46.0          & 61.7\,\raisebox{0.6ex}{\scriptsize\textcolor{green!55!black}{$+1.9$}}          & 57.4\,\raisebox{0.6ex}{\scriptsize\textcolor{green!55!black}{$+6.7$}}           & 40.1\,\raisebox{0.6ex}{\scriptsize\textcolor{red}{$-0.7$}}                     & 42.8\,\raisebox{0.6ex}{\scriptsize\textcolor{red}{$-0.2$}}                     & 46.9\,\raisebox{0.6ex}{\scriptsize\textcolor{green!55!black}{$+4.2$}}           & 41.1\,\raisebox{0.6ex}{\scriptsize\textcolor{green!55!black}{$+2.5$}}          & 48.4\,\raisebox{0.6ex}{\scriptsize\textcolor{green!55!black}{$+2.4$}} \\
\bottomrule
\end{tabular}%
}
\end{table*}

Table~\ref{tab:capability-main} shows that even the strongest models remain limited on complete-thread social reasoning.
Claude-Opus-$4.7$ performs best under the full-context Baseline, with a C1--C6 macro average of $60.0\%$.
Standard chain-of-thought prompting raises its average to $66.7\%$, while the C1--C6 macro average across all $18$ models increases modestly from $46.0\%$ to $48.4\%$.
For reference, five independent human readers average $72.4\%$ when answering from the complete discussions.

Overall, the strongest models perform better, while chain-of-thought provides only a modest aggregate gain and does not close the gap to human readers.

\subsection{Failures emerge as reasoning spans the thread}
\label{sec:results-capability}

Table~\ref{tab:capability-main} reveals a clear progression in difficulty.
Under the full-context Baseline, average accuracy is $59.8\%$ on \textbf{C1 local social meaning} and drops to $50.7\%$ on \textbf{C2 reply-pair interaction}.
For \textbf{C3--C6}, which require reasoning over larger discussion structures, accuracy falls further to $38.6\%$--$43.0\%$.
The sharpest failures appear on \textbf{C6 cross-subdiscussion variation} at $38.6\%$ and \textbf{C3 collective structure} at $40.8\%$, where models must compare behavior across branches or recover the positions organizing a discussion region.
The main difficulty therefore emerges when evidence must be integrated across the thread into a higher-level social judgment.

Standard chain-of-thought prompting does not remove this weakness.
It improves \textbf{C2} by $6.7$ pp, but leaves \textbf{C3 collective structure} and \textbf{C4 temporal dynamics} essentially unchanged, from $40.8\%$ to $40.1\%$ and from $43.0\%$ to $42.8\%$.
Cross-subdiscussion reasoning also remains low at $41.1\%$.
Thus, additional reasoning helps some local interaction decisions but does little to resolve the capabilities that require organizing evidence across larger thread structures, a pattern that also appears across individual models and platforms in Appendix~\ref{app:capability-platform}, Figs.~\ref{fig:capability-landscape-app}--\ref{fig:domain-method-app}.

\subsection{Reasoning benefits are strongly model-dependent}
\label{sec:results-cot}

Chain-of-thought prompting produces substantial gains for some models, but its effects vary considerably across the model pool.
Gemini-2.5-Pro and Claude-Sonnet-$4.6$ improve by $9.9$ and $8.3$ pp, respectively, while several other models show little improvement or decline under CoT.

This heterogeneity limits the aggregate benefit of additional reasoning.
Across all $18$ models, CoT raises the C1--C6 macro average by only $2.4$ pp, despite much larger gains for several individual models.
The results therefore show that the benefit of additional reasoning depends strongly on the model.
Appendix~\ref{app:reasoning-lift}, Table~\ref{tab:reasoning-lift} and Fig.~\ref{fig:reasoning-heatmap-app}, show that this model dependence extends beyond standard CoT to evidence enumeration, temporal tracking, option elimination, and self-reflection.

\subsection{Flat retrieval loses both evidence and thread structure}
\label{sec:results-retrieval}

\begin{figure*}[!t]
  \centering
  \includegraphics[width=0.92\linewidth]{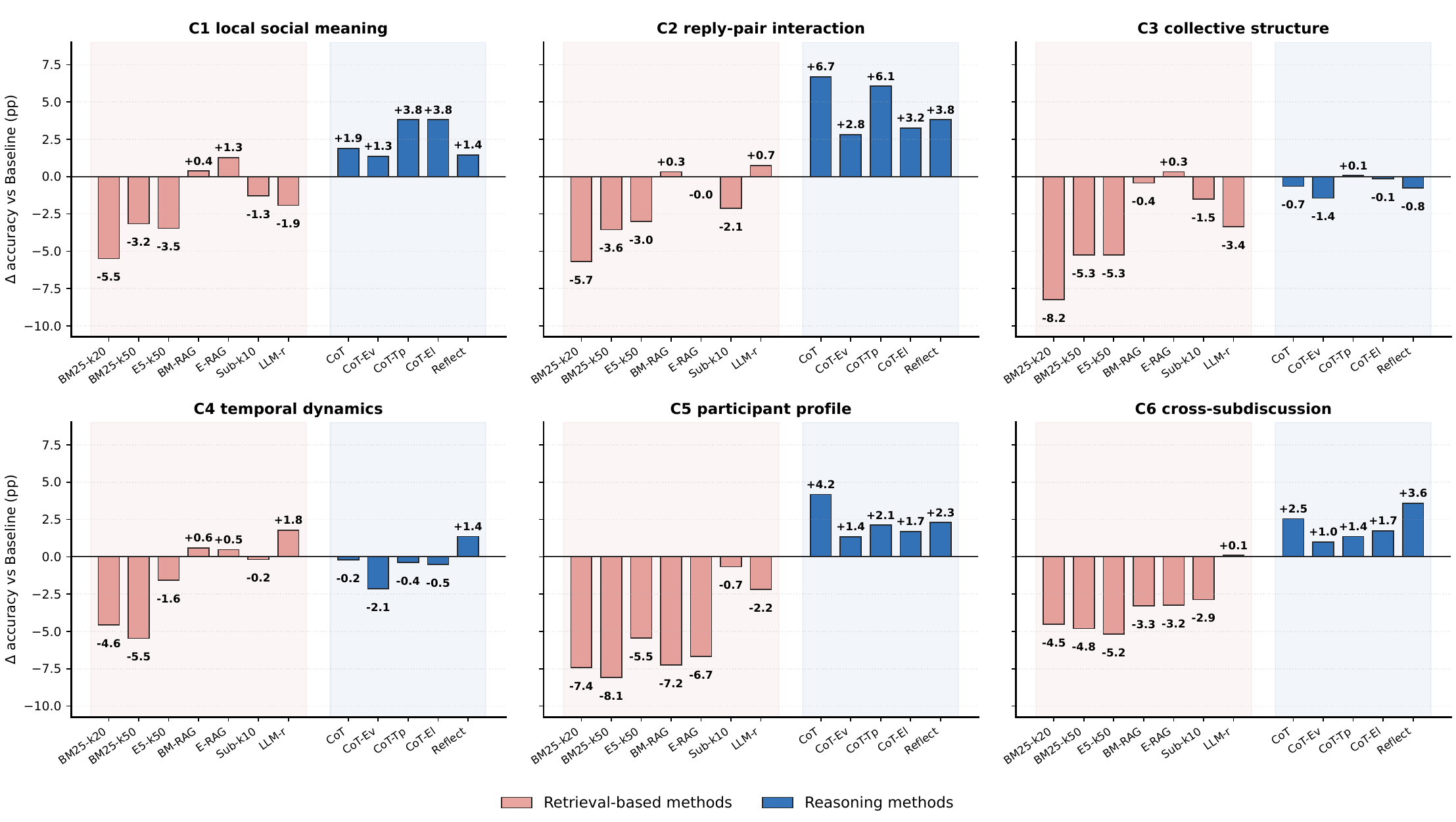}
  \caption{\textbf{Method effect by capability.}
    Mean $\Delta$ accuracy relative to Baseline across $18$ models for C1--C6.
    Reasoning methods are generally beneficial, while flat retrieval often reduces accuracy, especially on C3.
    Structure-aware and LLM-based retrieval recover part of these losses.}
  \label{fig:method-capability}
\end{figure*}

Figure~\ref{fig:method-capability} shows that standard comment-level retrieval is often counterproductive.
Both lexical BM$25$ and dense E$5$ retrieval reduce accuracy on several capabilities, with the largest losses on \textbf{C3 collective structure}.
Methods that preserve reply structure or let an LLM select coherent regions perform better, although they still provide only limited gains over the full-context Baseline.
The pattern suggests that relevance at the level of individual comments is a poor match for questions whose evidence is distributed across connected parts of the thread.

The evidence-coverage diagnostic reveals a complementary failure.
Standard comment-level retrievers recover only a small fraction of the author-cited evidence, with high-coverage retrieval remaining rare even as more comments are retrieved.
Thus, flat retrieval can fail before the final reasoning step by omitting necessary evidence, and it can further weaken the remaining evidence by separating comments from the reply structure in which they are meaningful.
Appendix~\ref{app:retrieval-diagnostics}, Table~\ref{tab:evidence-recall-app}, quantifies this gap through evidence recall at different retrieval budgets and the frequency of high-coverage retrieval.

\subsection{Validity checks support genuine thread-level difficulty}
\label{sec:results-validity}

We first test whether simpler cues can substitute for the full discussion.
The options-only, question-only, and title-and-root-only probes retain only the answer choices, the question stem, or the discussion title and root post, respectively.
Their accuracies remain limited at $31.5\%$, $38.0\%$, and $35.4\%$.
Retaining all comments but shuffling their order yields $40.4\%$.
Together, these probes show that answer-choice cues, question wording, and topic-level information are insufficient, while the organization of the discussion also contributes to successful reasoning.
Appendix~\ref{app:validity}, Table~\ref{tab:shortcut-robustness}, gives the full shortcut analysis.

We next examine whether the multiple-choice format drives the model comparison.
In the open-answer probe, models receive the same discussion and question without the answer choices and instead produce a short free-text response.
Accuracy drops by $7.6$ pp on the same items, but the model ranking is preserved almost exactly, with Spearman $\rho{=}1.00$.
Answer choices therefore make the task easier in absolute terms without materially changing which models perform better.

Finally, we test whether the remaining difficulty is primarily explained by contamination or input length.
A time-disjoint comparison finds no systematic advantage for the older Stack Exchange cohort over the 2026 Hacker News cohort across model tiers, providing no indication that source-page memorization is a major driver.
Separately, episode length has only a weak negative association with Baseline accuracy, with Spearman $\rho{=}{-}0.18$.
Longer discussions are somewhat harder, but length accounts for little of the observed performance variation.
Neither analysis supports memorization or episode length as a primary explanation for the benchmark difficulty.
Appendix~\ref{app:contamination-split}, Table~\ref{tab:contamination-app}, reports the time-disjoint comparison, while Appendix~\ref{app:length-effect}, Fig.~\ref{fig:length-vs-accuracy-app}, shows the episode-level length relationship.

\subsection{Failures arise in both evidence localization and criterion application}
\label{sec:results-component}

We next ask where the remaining errors arise.
Providing the gold evidence scope improves every model tier by roughly $10$--$12$ pp in Table~\ref{tab:goldscope-main}, raising the overall average from $43.9\%$ to $55.0\%$.
Evidence localization is therefore a substantial source of error, yet substantial difficulty remains even when the relevant thread region is known.

\begin{table}[t]
\small
\centering
\renewcommand{\arraystretch}{1.05}
\setlength{\tabcolsep}{5pt}
\caption{\textbf{Gold evidence scope improves accuracy across all model tiers, but substantial error remains.}
Mean accuracy across models in each tier on all $1{,}462$ items.}
\label{tab:goldscope-main}
\rowcolors{2}{rowshade}{white}
\begin{tabular}{@{}l c c c@{}}
\toprule
\rowcolor{white}
\textbf{Model tier} & \textbf{Full}   & \textbf{Gold-Scope} & \textbf{$\Delta$} \\
\midrule
Closed frontier (6) & $51.4$          & $62.3$              & $+10.9$ \\
Strong open (4)     & $43.4$          & $53.9$              & $+10.5$ \\
Mid open (4)        & $40.3$          & $51.2$              & $+10.9$ \\
Small (4)           & $37.0$          & $48.9$              & $+11.9$ \\
\midrule
\rowcolor{accentcell}
\textbf{All 18}     & \textbf{$43.9$} & \textbf{$55.0$}     & $\mathbf{+11.1}$ \\
\bottomrule
\end{tabular}
\end{table}

To understand what remains after evidence localization, we run a separate component oracle on Claude-Opus-$4.7$ using a stratified subset of $510$ items from $30$ episodes.
The CoT baseline reaches $63.1\%$.
Gold scope raises accuracy to $66.7\%$, while gold per-comment criterion labels raise it to $81.4\%$.
Providing both raises accuracy to $98.6\%$.
On this subset, the much larger improvement from criterion labels shows that knowing where to look is not enough: models must also determine correctly how each comment should be treated under the question-specific criterion.
Appendix~\ref{app:component-diagnostics}, Table~\ref{tab:component-oracle-app}, gives the full component comparison, including the model-generated enumerate-label-aggregate pipeline.

A manual audit of $50$ Claude-Opus-$4.7$ Gold-Scope errors supports the same diagnosis.
In $60\%$ of the audited errors, the model sees the relevant thread region but applies the question-specific criterion too broadly, too narrowly, or otherwise incorrectly.
The remaining errors involve thread structure, counting rules, or surface cues, with the full breakdown reported in Appendix~\ref{app:goldscope-errors}.

Together, these results identify two distinct sources of failure.
Models must first locate the relevant part of the discussion and then apply the correct criterion to the evidence within it.
Current reasoning and retrieval methods can improve parts of this process, but reliable performance requires both stages to succeed.
\section{Conclusion}
\label{sec:conclusion}

We introduced LongSocialBench, a benchmark of $1{,}462$ verified human-authored items over $94$ complete reply-structured discussions.
Across $18$ models and $29$ evaluation settings, current long-context workflows remain substantially below human readers, while additional reasoning and retrieval provide only partial and model-dependent improvements.
Our diagnostics identify two distinct sources of error: models must first locate the relevant part of the discussion and then correctly apply the question-specific social criterion to the evidence within it.
Even when the relevant region is supplied, criterion application remains a major source of failure.

These findings frame complete-thread social understanding as a structured reading problem over reply trees.
LongSocialBench provides a diagnostic testbed for developing models that can reliably reason across reply chains, participant trajectories, and cross-branch comparisons.

% \clearpage
\section*{Limitations}
\label{sec:limitations}

\paragraph{Scope of discussion data.}
LongSocialBench focuses on English, text-based, public reply-threaded discussions from Hacker News, Reddit r/ChangeMyView, and Stack Exchange.
These sources provide complete discussions with explicit reply structure, but they do not represent private or synchronous conversations, multimodal interaction, non-English discourse, or platforms with different interaction norms.
Future work should examine whether the capability surfaces and failure patterns identified here generalize to these broader conversational settings.

\paragraph{Benchmark format.}
LongSocialBench uses multiple-choice questions to support verifiable and reproducible evaluation of long-thread social reasoning.
Our open-answer probe shows that removing the answer choices lowers absolute accuracy while largely preserving the model ranking, suggesting that the main comparison is not specific to the multiple-choice format.
The benchmark nevertheless focuses on evidence-grounded social judgments and does not directly evaluate free-form explanation quality, calibrated uncertainty, or faithful long-thread summarization.
Future work can extend the benchmark with richer response formats and corresponding evaluation protocols.

\section*{Ethical Considerations}
\label{sec:ethics}

LongSocialBench is constructed from public discussions on Hacker News, Reddit r/ChangeMyView, and Stack Exchange under the applicable platform and licensing terms.
To reduce privacy risk, released discussions replace usernames with per-episode pseudonyms and remove personally identifying spans surfaced during verification.
Human contributors and readers follow a written annotation protocol and are compensated in accordance with institutional norms.
The benchmark evaluates observable interaction within public discussions and is not intended for identity inference, off-platform attribution, or individual-level profiling.
Further details on data handling, contributor procedures, and intended use are provided in Appendix~\ref{app:ethics}.

\section*{Acknowledgments}

We thank the contributors and human readers who supported the construction and evaluation of LongSocialBench. This work was supported in part by the Defense Advanced Research Projects Agency (DARPA) under the Methodological Advancements for Generalizable Insights into Complex Systems (MAGICS) program, Agreement No. HR0011263E079.

\bibliographystyle{acl_natbib}
\bibliography{bib/refs}

\appendix

\noindent\textbf{Appendix overview.}
Appendix~\ref{app:dataset-coverage} characterizes the discussion corpus, while Appendices~\ref{app:item-design} and~\ref{app:verification-details} detail item construction, safeguards, and verification.
Appendix~\ref{app:methods} specifies the evaluated models and workflows, and Appendix~\ref{app:detailed-results} provides the additional analyses supporting the main results.
Appendix~\ref{app:samples} gives representative questions across the six capability surfaces.
Finally, Appendices~\ref{app:portal}, \ref{app:ethics}, and~\ref{app:reproducibility} document the construction protocol, ethical and licensing considerations, and release and reproducibility details.

\section{Dataset Coverage and Episode Statistics}
\label{app:dataset-coverage}

This appendix section supports the episode-sampling claims in \S\ref{sec:data-collection}. 
It reports the released snapshot, topic coverage, episode scale, and structural diversity of LongSocialBench.

\subsection{Topic Coverage}
\label{app:topic-coverage}

Table~\ref{tab:topic-coverage} summarizes the primary-topic coverage of the $94$ released episodes. 
Because the three platforms do not share a single topic taxonomy, each episode is assigned one broad topic family based on its title and, for Stack Exchange, its site domain.
These labels summarize dataset coverage and are not used as model-facing metadata or evaluation labels.

\subsection{Episode Scale and Structural Diversity}
\label{app:episode-statistics}

\begin{figure*}[!htbp]
\centering
\includegraphics[width=\linewidth]{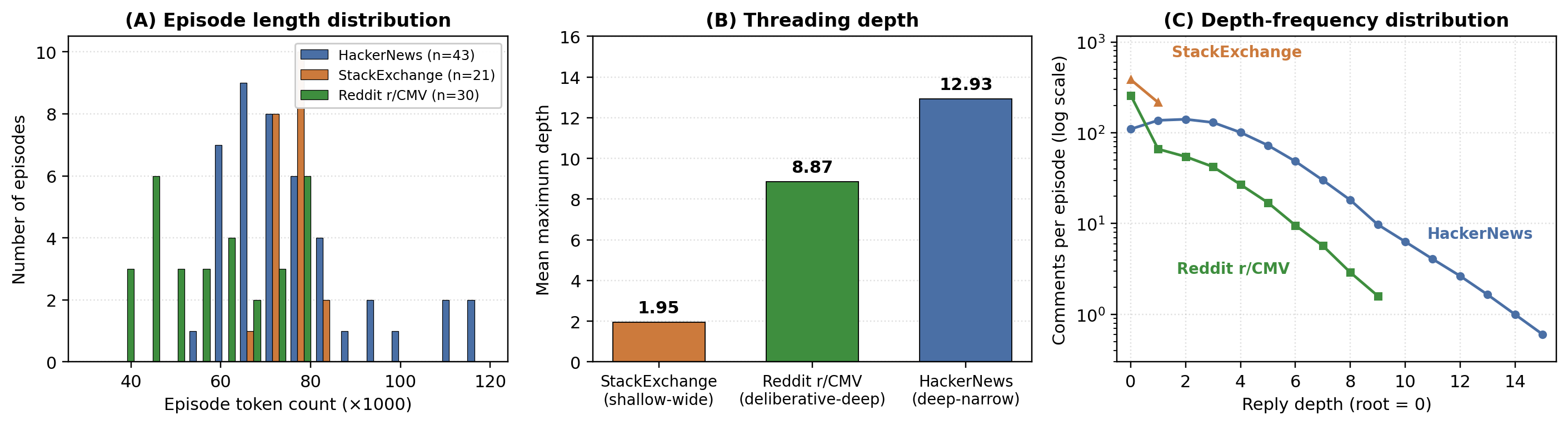}
\caption{\textbf{Structural diversity of LongSocialBench episodes.}
\textbf{(A)} Episode lengths span $37$ to $119$K tokens with platform-specific medians of $73$K for Hacker News, $76$K for Stack Exchange, and $56$K for Reddit r/CMV.
\textbf{(B)} Mean maximum depth differs substantially across sources: Stack Exchange $2.0$, Reddit r/CMV $8.9$, and Hacker News $12.9$.
\textbf{(C)} Depth-frequency curves show different threading regimes: Stack Exchange is concentrated near depths $1$ to $2$, Reddit has more mass at depths $1$ to $6$, and Hacker News retains substantial mass at depths $4$ to $10$.}
\label{fig:dataset-overview}
\end{figure*}

Table~\ref{tab:episode-stats} reports the scale and structural profile of the released episodes by source.
The three sources provide distinct discussion shapes: Hacker News has the deepest reply chains on average, Stack Exchange is shallow but wide, and Reddit r/ChangeMyView contains deliberative multi-step exchanges.
Fig.~\ref{fig:dataset-overview} visualizes the same coverage along serialized length, maximum depth, and platform-specific depth-frequency curves.

\section{Item Design and Construction Safeguards}
\label{app:item-design}

This appendix expands the capability surfaces introduced in \S\ref{sec:task-overview} and documents the release criteria and shortcut controls used during item construction, complementing the data-collection details in \S\ref{sec:data-collection}.

\subsection{Capability Surfaces in Detail}
\label{app:capability-table}

Table~\ref{tab:capability-design-app} expands the six capability surfaces introduced in \S\ref{sec:task-overview}.
The table summarizes what each surface asks the model to decide and illustrates it with abridged examples adapted from human-authored items.

\begin{table*}[!htbp]
\small
\centering
\renewcommand{\arraystretch}{1.12}
\setlength{\tabcolsep}{5pt}
\caption{\textbf{Coarse topic coverage of released episodes.}
Each episode is assigned one primary topic family based on its title and source domain.}
\label{tab:topic-coverage}
\rowcolors{2}{rowshade}{white}
\begin{tabular}{@{}lrrrr@{}}
\toprule
\rowcolor{white}
\textbf{Primary topic family} 
& \textbf{Total} 
& \textbf{HN} 
& \textbf{Reddit} 
& \textbf{SE} \\
\midrule
Computing / AI / internet        & 25          & 20          & 0           & 5 \\
Economy / labor / business       & 21          & 10          & 10          & 1 \\
Science / engineering / health   & 17          & 3           & 5           & 9 \\
Public policy / law / governance & 11          & 6           & 4           & 1 \\
Culture / media / social life    & 7           & 3           & 4           & 0 \\
Education / academia             & 6           & 0           & 4           & 2 \\
Climate / energy / environment   & 4           & 1           & 3           & 0 \\
Humanities / social thought      & 3           & 0           & 0           & 3 \\
\midrule
\rowcolor{accentcell}
\textbf{Total}                   & \textbf{94} & \textbf{43} & \textbf{30} & \textbf{21} \\
\bottomrule
\end{tabular}
\end{table*}

\begin{table*}[!htbp]
\centering
\renewcommand{\arraystretch}{1.16}
\setlength{\tabcolsep}{6pt}
\caption{\textbf{Episode-level scale and structure by source.}
Token counts use \texttt{cl100k\_base} over the complete serialized discussion.
Maximum depth is measured from the root, and root replies are comments at depth $1$.}
\label{tab:episode-stats}
\rowcolors{2}{rowshade}{white}
\resizebox{\linewidth}{!}{%
\begin{tabular}{@{}l r rrr rrr@{}}
\toprule
\rowcolor{white}
\textbf{Source}  & \textbf{\# eps} & \textbf{Tokens med.}  & \textbf{Tokens max} & \textbf{Tokens total}   & \textbf{Mean max-depth} & \textbf{Mean nodes} & \textbf{Mean root replies} \\
\midrule
Hacker News      & 43              & 73{,}397              & 118{,}981           & 3.29M                   & 12.9                    & 815.7               & 109.9 \\
Stack Exchange   & 21              & 75{,}982              & 85{,}184            & 1.59M                   & 2.0                     & 601.0               & 385.2 \\
Reddit r/CMV     & 30              & 56{,}322              & 79{,}597            & 1.75M                   & 8.9                     & 485.5               & 258.8 \\
\midrule
\rowcolor{accentcell}
\textbf{Overall} & \textbf{94}     & \textbf{$\approx$73K} & \textbf{118{,}981}  & \textbf{$\approx$6.63M} & \textbf{9.2}            & \textbf{662.4}      & \textbf{218.9} \\
\bottomrule
\end{tabular}}
\end{table*}

\begin{table*}[!htbp]
\small
\centering
\renewcommand{\arraystretch}{1.12}
\setlength{\tabcolsep}{5pt}
\caption{\textbf{Six capability surfaces in LongSocialBench.}
Each surface is a recurring social-reading decision over a complete discussion episode.}
\label{tab:capability-design-app}
\rowcolors{2}{rowshade}{white}
\begin{tabularx}{\linewidth}{@{}
>{\raggedright\arraybackslash}p{2.75cm}
>{\raggedright\arraybackslash}p{3.35cm}
>{\raggedright\arraybackslash}X
@{}}
\toprule
\rowcolor{white}
\textbf{Capability surface} 
& \textbf{Decision tested} 
& \textbf{Abridged example item} \\
\midrule

\textbf{C1: Local social meaning}
& Decide what one comment is doing in context.
& A target comment replies to a parent claim. Decide its \textbf{local role}: support, pushback, reframing, or clarification. \\

\textbf{C2: Reply-pair interaction}
& Compare a reply with the comment it answers.
& A parent uses one basis of comparison; the reply switches basis. Decide the \textbf{parent reply relation}: refute, qualify, redirect, or concede. \\

\textbf{C3: Collective structure}
& Identify the positions organizing a scoped discussion region.
& In a scoped subtree, choose the \textbf{position structure} that best organizes the replies: framing, consequences, responsibility, or evidence. \\

\textbf{C4: Temporal dynamics}
& Locate where an ordered exchange changes direction.
& In a long reply chain, identify the first \textbf{turning point} where one argumentative framing gives way to another. \\

\textbf{C5: Participant profile}
& Track one participant across comments.
& Across one author's comments, infer the \textbf{participant trajectory}: stable stance, narrowed claim, concession, escalation, or strategy shift. \\

\textbf{C6: Cross-subdiscussion variation}
& Compare behavior across separate branches.
& Across two branches, decide the \textbf{cross-branch change}: whether a role, stance, or strategy stays stable or shifts. \\

\bottomrule
\end{tabularx}
\end{table*}

\subsection{Release Criteria}
\label{app:release-criteria-table}

Table~\ref{tab:release-criteria-app} summarizes the criteria that every candidate item must satisfy before release.

\begin{table*}[!htbp]
\small
\centering
\renewcommand{\arraystretch}{1.07}
\setlength{\tabcolsep}{5pt}
\caption{\textbf{Release criteria for LongSocialBench items.}
Each candidate item must satisfy all criteria before release; failed items are clarified and re-verified, or removed if the issue cannot be resolved.}
\label{tab:release-criteria-app}
\rowcolors{2}{rowshade}{white}
\begin{tabularx}{\linewidth}{@{}p{2.55cm} X X@{}}
\toprule
\rowcolor{white}
\textbf{Criterion} & \textbf{Requirement} & \textbf{Failure action} \\
\midrule

Answerability
& The gold option is recoverable from the supplied episode alone, without external facts or author intent.
& Rewrite or remove. \\

Option uniqueness
& Exactly one option is supported under the stated scope and author-written criterion.
& Revise options or remove. \\

Scope clarity
& The stem specifies the relevant unit, such as a comment, direct-reply set, subtree, chain, participant trajectory, or cross-branch comparison.
& Clarify stem and re-verify. \\

Evidence sufficiency
& Evidence nodes support both the gold option and key premises such as branch boundary, counting unit, participant trajectory, or comparison set.
& Add evidence, rewrite, or remove. \\

Observable social criterion
& Interpretive questions are grounded in observable thread behavior rather than preference, speculation, or unstated norms.
& Rewrite criterion or remove. \\

Option balance
& Distractors are plausible wrong readings with comparable specificity, not fillers or wording artifacts.
& Rewrite options and re-verify. \\

\bottomrule
\end{tabularx}
\end{table*}

\subsection{Shortcut and Contamination Controls}
\label{app:anti-shortcut-controls}

Table~\ref{tab:pillars-summary-app} summarizes the authoring controls used to reduce surface shortcuts, improve distractor quality, and screen public-source contamination.

\begin{table*}[!htbp]
\small
\centering
\renewcommand{\arraystretch}{1.06}
\setlength{\tabcolsep}{5pt}
\caption{\textbf{Question-design controls used during authoring.}
The controls support shortcut-resistant multiple-choice evaluation over complete discussion episodes.}
\label{tab:pillars-summary-app}
\rowcolors{2}{rowshade}{white}
\begin{tabularx}{\linewidth}{@{}p{2.45cm} X X@{}}
\toprule
\rowcolor{white}
\textbf{Control} & \textbf{Shortcut reduced} & \textbf{Authoring restriction} \\
\midrule

Minimal locators
& The stem reveals the relevant node, chain, or subtree.
& Use only the locator needed to define the scope; avoid exact quotes, usernames, redundant structural hints, or answer-implying descriptors. \\

Distributed evidence
& A single salient comment solves a thread-level item.
& Require evidence from the intended scope whenever the question asks for aggregation, comparison, trajectory, or cross-branch reasoning. \\

Controlled options
& Option wording makes the answer obvious.
& Use parallel option forms and close numeric or relational alternatives when appropriate; avoid answer-specific descriptive phrasing. \\

No hint vocabulary
& Keyword search or hint words reveal the answer.
& State the criterion without listing answer-bearing instances or repeating distinctive phrases from the gold evidence. \\

Diagnostic distractors
& Distractors are implausible or too easy.
& Tie distractors to plausible wrong scopes, wrong criteria, lexical matches, boundary errors, or aggregation mistakes. \\

Open-web contamination
& The item is recoverable from open-web memory rather than the supplied episode.
& Check distinctive stem/option phrases and named events; rewrite or exclude items whose answers can be recovered from open-web summaries or third-party transcripts. \\

\bottomrule
\end{tabularx}
\end{table*}

\section{Verification and Release Details}
\label{app:verification-details}

This appendix section supports the verification and release claims in \S\ref{sec:verification}.
It records the independent verification protocol and release audit.

\subsection{Verification Protocol}
\label{app:verification-protocol}

Every released LongSocialBench item passes independent verification or adjudication.
In Phase~1, the author writes the question stem, answer options, gold option, plain-language criterion, and evidence nodes.
In Phase~2, a different trained contributor answers from the same model-facing input used in evaluation: the serialized thread, question stem, and four answer options, without the author gold option, criterion note, or evidence-node list.
In Phase~3, verifier disagreements and author-flagged edge cases are checked by a third reviewer, who confirms the item, requires revision and a new Phase-2 verification pass, or removes it.

Among the $1{,}553$ parseable Phase-2 candidate records, independent verification matches the author gold option in $1{,}419$ cases, yielding a candidate-level first-pass agreement of \textbf{$91.37\%$} before adjudication and Cohen's $\kappa{=}\mathbf{0.88}$ against a four-option chance baseline ($P_e{=}0.25$).
Of the $134$ first-pass mismatches, $43$ are resolved by adjudication and retained, while $91$ unresolved cases are excluded.
Phase~3 also reviews $24$ author-flagged edge cases that had passed first-pass agreement.
The $43$ retained mismatch cases and $24$ author-flagged cases yield $67$ Phase-3 records carried forward in the release, with verdicts of $21$ confirm, $40$ clarify, and $6$ rewrite.

\subsection{Stratified Release Audit}
\label{app:release-audit}

The release audit estimates residual item-level errors after verification and adjudication.
We sample $100$ released items stratified by capability surface and platform, with allocation approximately proportional to the benchmark distribution and coverage of every non-empty capability--platform combination.

For each item, the audit checks that the cited evidence resolves, the criterion matches the intended scope, the stem and options are well formed, and no infrastructure or option-design artifacts reveal the answer.
An adjudicator then re-reads the item from the model-facing input and assigns one of three verdicts.
\textsc{Pass} requires no correction, \textsc{corrected} indicates a wording-level revision that leaves the gold option unchanged, and \textsc{residual} indicates that the gold option cannot be recovered from the supplied input.

\begin{table}[!t]
\small
\centering
\renewcommand{\arraystretch}{1.08}
\setlength{\tabcolsep}{6pt}
\caption{\textbf{Stratified release audit.}
Among $100$ audited items, $98$ require no correction, two require wording-level revisions without changing the gold option, and none contains a residual gold-answer error.}
\label{tab:release-audit}
\rowcolors{2}{rowshade}{white}
\begin{tabular}{@{}l c c c@{}}
\toprule
\rowcolor{white}
\textbf{Domain} & \textbf{Pass}     & \textbf{Corrected} & \textbf{Residual} \\
\midrule
Hacker News      & 44 / 46           & 2 / 46             & 0 / 46 \\
Reddit r/CMV     & 32 / 32           & 0 / 32             & 0 / 32 \\
Stack Exchange   & 22 / 22           & 0 / 22             & 0 / 22 \\
\midrule
\rowcolor{accentcell}
\textbf{All}     & \textbf{98 / 100} & \textbf{2 / 100}   & \textbf{0 / 100} \\
\bottomrule
\end{tabular}\\[3pt]
{\footnotesize No residual gold-option errors were found in any capability. The two \textsc{corrected} items occurred in C2 and C3, one each.}
\end{table}
\section{Model and Workflow Details}
\label{app:methods}

This appendix provides the model list and workflow settings referenced in \S\ref{sec:setup}.
It complements the main text by specifying the evaluated model set, evaluation context caps, workflow construction, and workflow-specific prompts.

\subsection{Models}
\label{app:model-snapshots}

Table~\ref{tab:model-snapshots} summarizes the models used in the evaluation grid.
The set covers closed-source frontier systems and open-weight systems across large, mid-tier, and smaller model families.
We report the evaluation context cap because full-context workflows require the serialized episode prompt to fit within the window configured for the run.

\begin{table}[!t]
\small
\centering
\renewcommand{\arraystretch}{1.08}
\setlength{\tabcolsep}{5pt}
\caption{\textbf{Models used in the evaluation grid.}
CW = evaluation context cap in tokens.}
\label{tab:model-snapshots}
\rowcolors{2}{rowshade}{white}
\begin{tabular}{@{}l l r@{}}
\toprule
\rowcolor{white}
\textbf{Model}       & \textbf{Source} & \textbf{CW} \\
\midrule
\multicolumn{3}{@{}l}{\textit{Closed-source frontier systems}}\\
Claude-Opus-4.7      & Anthropic       & 200K \\
GPT-5                & OpenAI          & 400K \\
Claude-Sonnet-4.5    & Anthropic       & 200K \\
Claude-4             & Anthropic       & 200K \\
Gemini-2.5-Pro       & Google          & 1M \\
Claude-Sonnet-4.6    & Anthropic       & 200K \\
\midrule
\multicolumn{3}{@{}l}{\textit{Open-weight systems}}\\
Mistral-Large-3      & HuggingFace     & 128K \\
Llama-4-Scout-17B    & HuggingFace     & 128K \\
GLM-5                & HuggingFace     & 128K \\
Qwen3-Next-80B       & HuggingFace     & 256K \\
Llama-3.1-70B        & HuggingFace     & 128K \\
DeepSeek-R1          & HuggingFace     & 128K \\
Qwen3-32B            & HuggingFace     & 128K \\
Gemma-3-27B          & HuggingFace     & 128K \\
GPT-OSS-120B         & HuggingFace     & 128K \\
Llama-4-Maverick-17B & HuggingFace     & 128K \\
Llama-3.2-90B        & HuggingFace     & 128K \\
GPT-OSS-20B          & HuggingFace     & 128K \\
\bottomrule
\end{tabular}
\end{table}

\subsection{Workflow Settings and Prompts}
\label{app:workflow-settings}

Table~\ref{tab:workflow-settings} summarizes the $29$ evaluation settings used in the paper, grouped as $22$ deployable workflows (Baseline, CoT variants, retrieval variants) and $7$ diagnostic settings (validity probes, scope diagnostics).
The table records how each setting constructs the model input and guides the answer.
Deployable workflows and validity probes exclude author-side gold labels, criteria, evidence nodes, verifier notes, and adjudication records from prompting.
Scope diagnostics additionally provide either gold or inferred scope information, as specified below.
For multiple-choice settings, responses use the same deterministic answer parser and exact-match scoring rule; open-answer responses are content-scored separately.
Per-response parse status is retained for reproducibility.

\begin{table*}[!htbp]
\footnotesize
\centering
\renewcommand{\arraystretch}{1.06}
\setlength{\tabcolsep}{4pt}
\caption{\textbf{Evaluation settings on LongSocialBench.}
Deployable workflows are used for the headline accuracy comparison. Diagnostic settings test shortcut artifacts, multiple-choice format effects, or evidence localization and are reported separately.}
\label{tab:workflow-settings}
\rowcolors{2}{rowshade}{white}
\begin{tabularx}{\linewidth}{@{}
>{\raggedright\arraybackslash}p{2.35cm}
>{\raggedright\arraybackslash}p{3.05cm}
>{\raggedright\arraybackslash}X
@{}}
\toprule
\rowcolor{white}
\textbf{Setting group}
& \textbf{Variants}
& \textbf{Input construction and prompt} \\
\midrule

\multicolumn{3}{@{}l}{\textit{Deployable workflows}}\\

Full-context prompting
& Full
& The model receives the complete serialized discussion, reply-structure information, question stem, and four options; the prompt asks it to locate relevant evidence, compare options, and output one answer letter. \\

Reasoning prompts
& CoT, CoT-Ev, CoT-Tp, CoT-El, Reflect
& Same input as Full; the prompt adds one reasoning strategy before the final answer: free-form reasoning, evidence enumeration, temporal tracking, option elimination, or self-reflection. \\

Sparse retrieval
& BM25-$k20$, BM25-$k50$
& Comments are ranked by BM25 using the question stem as query; the top-$k$ comments are passed in original thread order and answered with the shared answer format. \\

Dense retrieval
& E5-$k20$, E5-$k50$
& Comments and question stems are embedded with E5; the top-$k$ comments by cosine similarity are passed in original thread order and answered with the shared answer format. \\

RAG with anchor notes
& BM25/E5 $\times$ $k\in\{20,50\}$ $\times$ anchors $\in\{1,2\}$
& Retrieved comments are supplemented with brief model-generated anchor notes that identify what each retrieved region contributes before final answering. \\

Structure-aware retrieval
& Subtree-$k1$, Subtree-$k5$, Subtree-$k10$
& The model receives the largest connected reply subtrees by node count, preserving local reply structure without using gold evidence, and answers from those subtrees. \\

LLM-as-retriever
& LLM-r
& The model first selects relevant comments or reply regions; the final answer prompt uses only the selected context. The selection prompt asks for evidence needed to answer the question, not merely lexical overlap. \\

\midrule
\multicolumn{3}{@{}l}{\textit{Diagnostic settings}}\\

Validity probes
& Options-only, question-only, title/root-only, shuffled-thread, open-answer
& The model receives a reduced or perturbed input; for open-answer, the four options are removed and the response is content-scored against the gold answer content. These probes test answer-option artifacts, stem leakage, topic familiarity, thread-order dependence, and multiple-choice format effects. \\

Scope diagnostics
& Gold-scope, infer-scope
& The model receives diagnostic scope information, either from gold scope metadata or from a separate inferred-scope prompt. These settings estimate the role of evidence localization rather than serving as standard deployment workflows. \\

\bottomrule
\end{tabularx}
\end{table*}
\section{Additional Evaluation Results}
\label{app:detailed-results}

This appendix reports additional results in the same sequence as the main Results section: headline robustness and the human comparison, capability and platform profiles, reasoning prompts, retrieval diagnostics, validity checks, and component diagnostics.

\subsection{Headline Robustness}
\label{app:headline-robustness}

\begin{figure*}[!htbp]
\centering
\includegraphics[width=0.95\linewidth]{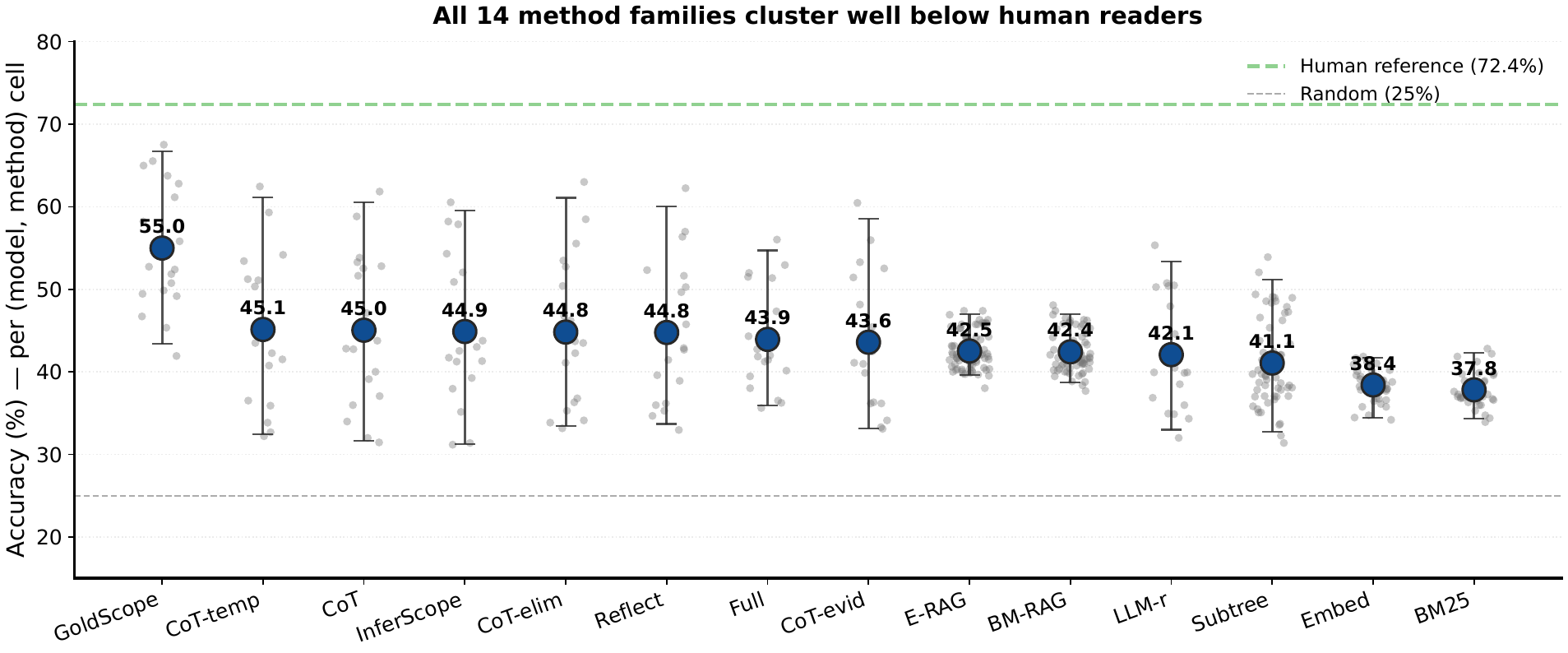}
\caption{\textbf{Method-family accuracy band on LongSocialBench.}
Each grey dot is one model-workflow cell from the $18 \times 29$ evaluation grid; markers and error bars show the family mean and $\pm 1$ standard deviation across cells.
All deployable workflows cluster below the human answerability reference of $72.4\%$; the Gold-Scope diagnostic reaches $55.0\%$.}
\label{fig:method-family-band-app}
\end{figure*}

Table~\ref{tab:method-family-ci-app} and Fig.~\ref{fig:method-family-band-app} summarize the accuracy band of the deployable workflows together with the two scope diagnostics for reference.
Table~\ref{tab:human-reference} reports the human answerability reference used in the main comparison.

\begin{table*}[!t]
\small
\centering
\renewcommand{\arraystretch}{1.05}
\setlength{\tabcolsep}{8pt}
\caption{\textbf{Per-model accuracy on three representative workflows.} Full = full-context prompting; CoT = chain-of-thought; LLM-r = LLM-as-retriever. Best$^\dagger$ is the per-model maximum across all evaluated non-oracle workflows, reported as a diagnostic only. C-mac = macro mean over the six capability surfaces.}
\label{tab:headline-results-app}
\begin{tabular}{@{}l rrrrr@{}}
\toprule
\textbf{Model}     & \textbf{Full} & \textbf{CoT} & \textbf{LLM-r} & \textbf{Best$^\dagger$} & \textbf{C-mac} \\
\midrule
\multicolumn{6}{@{}l}{\emph{(Closed-source frontier)}} \\
Claude-Opus-4.7    & 56.0          & 61.8         & 55.3           & 63.0                    & 60.0 \\
Gemini-2.5-Pro     & 51.5          & 58.8         & 50.4           & 59.3                    & 51.3 \\
GPT-5              & 52.9          & 53.8         & 50.3           & 56.4                    & 58.3 \\
Claude-Sonnet-4.5  & 51.4          & 52.8         & 50.8           & 53.4                    & 54.4 \\
Claude-4           & 52.0          & 52.5         & 50.5           & 53.5                    & 57.0 \\
Claude-Sonnet-4.6  & 44.3          & 51.6         & 39.9           & 51.6                    & 45.7 \\
\midrule
\multicolumn{6}{@{}l}{\emph{(Strong open-weight)}} \\
GLM-5              & 41.5          & 53.3         & 44.6           & 53.3                    & 41.3 \\
DeepSeek-R1        & 47.3          & 47.1         & 47.9           & 49.0                    & 50.6 \\
Qwen3-Next-80B     & 42.7          & 42.8         & 39.9           & 43.7                    & 44.6 \\
Mistral-Large-3    & 42.0          & 42.7         & 38.5           & 43.5                    & 43.3 \\
\midrule
\multicolumn{6}{@{}l}{\emph{(Mid open-weight)}} \\
Llama-4-Mav-17B    & 41.2          & 43.8         & 40.5           & 45.3                    & 41.5 \\
Llama-4-Scout-17B  & 40.2          & 40.0         & 36.9           & 41.7                    & 41.3 \\
Llama-3.1-70B      & 38.0          & 39.1         & 36.0           & 41.1                    & 39.5 \\
Qwen3-32B          & 41.9          & 31.5         & 34.9           & 43.3                    & 43.8 \\
\midrule
\multicolumn{6}{@{}l}{\emph{(Small / mid-weak)}} \\
GPT-OSS-120B       & 39.5          & 37.1         & 34.3           & 42.2                    & 42.0 \\
GPT-OSS-20B        & 36.3          & 36.0         & 32.0           & 40.2                    & 39.2 \\
Llama-3.2-90B      & 35.6          & 34.0         & 39.9           & 40.0                    & 35.3 \\
Gemma-3-27B        & 36.5          & 32.0         & 35.0           & 40.4                    & 38.1 \\
\midrule
\rowcolor{accentcell}
\textbf{Mean (18)} & 43.9          & 45.0         & 42.1           & 47.8                    & 46.0 \\
\bottomrule
\end{tabular}
\end{table*}

\begin{table*}[!htbp]
\small
\centering
\renewcommand{\arraystretch}{1.05}
\setlength{\tabcolsep}{8pt}
\caption{\textbf{Accuracy band by setting group, with cell-level bootstrap 95\% CIs.}
Values are full-benchmark micro accuracy on the $1{,}462$ released items, averaged over the per-cell accuracies inside each group.
$\Delta$Full is the lift over Full-context (Baseline). Cells $=$ number of (model, setting) cells in the group; the eight rows above the divider are deployable workflows, and the two rows below are scope diagnostics reported for reference.}
\label{tab:method-family-ci-app}
\rowcolors{2}{rowshade}{white}
\begin{tabular}{@{}l r c c c@{}}
\toprule
\rowcolor{white}
\textbf{Family}           & \textbf{Cells} & \textbf{Mean} & \textbf{95\% CI} & \textbf{$\Delta$Full} \\
\midrule
Chain-of-thought          & 90             & 44.7          & [43.0, 46.5]     & $+0.7$ \\
\rowcolor{accentcell}
Full-context              & 18             & 43.9          & [41.3, 46.9]     & --- \\
RAG $k{=}50$              & 72             & 42.7          & [42.1, 43.2]     & $-1.3$ \\
RAG $k{=}20$              & 72             & 42.3          & [41.8, 42.8]     & $-1.6$ \\
LLM-as-retriever          & 18             & 42.1          & [39.1, 45.4]     & $-1.8$ \\
Structure-aware retrieval & 54             & 41.1          & [39.6, 42.6]     & $-2.9$ \\
Retrieval $k{=}50$        & 36             & 39.0          & [38.4, 39.7]     & $-4.9$ \\
Retrieval $k{=}20$        & 36             & 37.2          & [36.6, 37.9]     & $-6.7$ \\
\midrule
Infer-Scope diagnostic    & 18             & 44.9          & [40.8, 49.2]     & $+1.0$ \\
Gold-Scope diagnostic     & 18             & 55.0          & [51.8, 58.4]     & $+11.1$ \\
\bottomrule
\end{tabular}
\end{table*}

\subsection{Human Reference and Matched Comparison}
\label{app:human-reference}

The human reference reported in the main text is collected from five graduate-student readers with no prior exposure to LongSocialBench.
They answer the same stratified $150$-item sample under the same information condition as models: serialized thread, reply structure, question stem, and four answer options, with no criterion notes, author evidence records, or verifier annotations.
The mean accuracy is $72.4\%$.
These five readers are a separate cohort from the construction-time verifier pool, which involved the full $20$-contributor team during Phase~2 cross-verification and reached $91.37\%$ candidate-level first-pass agreement (Cohen's $\kappa{=}0.88$) on the $1{,}553$ parseable Phase-2 candidate records before adjudication.
The five-reader results provide the human answerability and model-comparison reference.

\begin{table}[!htbp]
\small
\centering
\renewcommand{\arraystretch}{1.05}
\setlength{\tabcolsep}{5pt}
\caption{\textbf{Human answerability reference.}
Five readers answer the same stratified $150$-item sample without LongSocialBench-specific training.}
\label{tab:human-reference}
\rowcolors{2}{rowshade}{white}
\begin{tabular}{@{}l c c@{}}
\toprule
\rowcolor{white}
\textbf{Reader} & \textbf{Accuracy} & \textbf{\# correct / 150} \\
\midrule
R1              & 74.7              & 112 \\
R2              & 73.3              & 110 \\
R3              & 72.7              & 109 \\
R4              & 71.3              & 107 \\
R5              & 70.0              & 105 \\
\midrule
\rowcolor{accentcell}
\textbf{Mean}   & \textbf{72.4}     & \textbf{108.6} \\
\bottomrule
\end{tabular}\\[2pt]
{\footnotesize 95\% bootstrap CI on the reader mean: $[71.1, 73.7]$. Random multiple-choice baseline: 25.0\%.}
\end{table}

\subsubsection{Matched Model--Human Comparison}
\label{app:matched-aggregation}

The headline mean-vs-mean comparison computes the model number on the full benchmark and the human number on a stratified $150$-item subsample.
We therefore re-run the model evaluation on a stratified $150$-item subset whose capability-platform proportions match the human reference sample, bootstrap-resampled $1000$ times.
Table~\ref{tab:matched-app} reports model statistics matched to the human comparator.

\begin{table}[!htbp]
\small
\centering
\renewcommand{\arraystretch}{1.05}
\setlength{\tabcolsep}{4pt}
\caption{\textbf{Stratified $150$-item matched-aggregation subset.}
Bootstrap ($B{=}1000$) over draws of $150$ items in the natural capability-platform proportions.
Each row reports best-cell mean and the $95\%$ CI across draws; human-reader numbers are for reference.}
\label{tab:matched-app}
\rowcolors{2}{rowshade}{white}
\begin{tabular}{@{}l c c@{}}
\toprule
\rowcolor{white}
\textbf{Cohort}                & \textbf{Mean}   & \textbf{$95\%$ CI} \\
\midrule
Top single cell                & $66.1$          & $[61.1, 71.8]$ \\
Claude-Opus-$4.7$              & $65.4$          & $[59.7, 71.8]$ \\
Closed-frontier mean ($n{=}6$) & $59.3$          & --- \\
Strong-open mean ($n{=}4$)     & $51.1$          & --- \\
$18$-model fleet mean          & $51.6$          & --- \\
\midrule
\rowcolor{accentcell}
Human readers ($n{=}5$)        & $\mathbf{72.4}$ & $\mathbf{[71.1, 73.7]}$ \\
Worst human reader (R$5$)      & $70.0$          & --- \\
\bottomrule
\end{tabular}\\[2pt]
{\footnotesize Mean = best-cell mean across $1000$ bootstrap draws of $150$ stratified
items. Best-cell = each model's highest-accuracy deployable workflow on the draw.}
\end{table}

\subsection{Capability and Platform Profiles}
\label{app:capability-platform}

\begin{figure}[!t]
  \centering
  \includegraphics[width=0.99\linewidth]{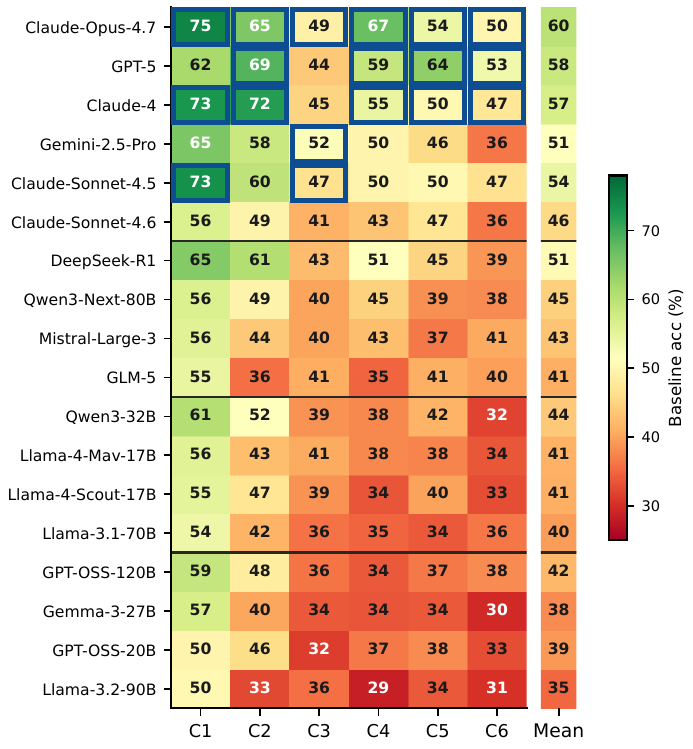}
  \caption{\textbf{Per-model accuracy on each social capability under Baseline.}
  Cells are mean accuracy on the $1{,}462$-item benchmark restricted to that capability; the rightmost strip is the per-model mean.}
  \label{fig:capability-landscape-app}
\end{figure}

\begin{figure}[!t]
  \centering
  \includegraphics[width=0.99\linewidth]{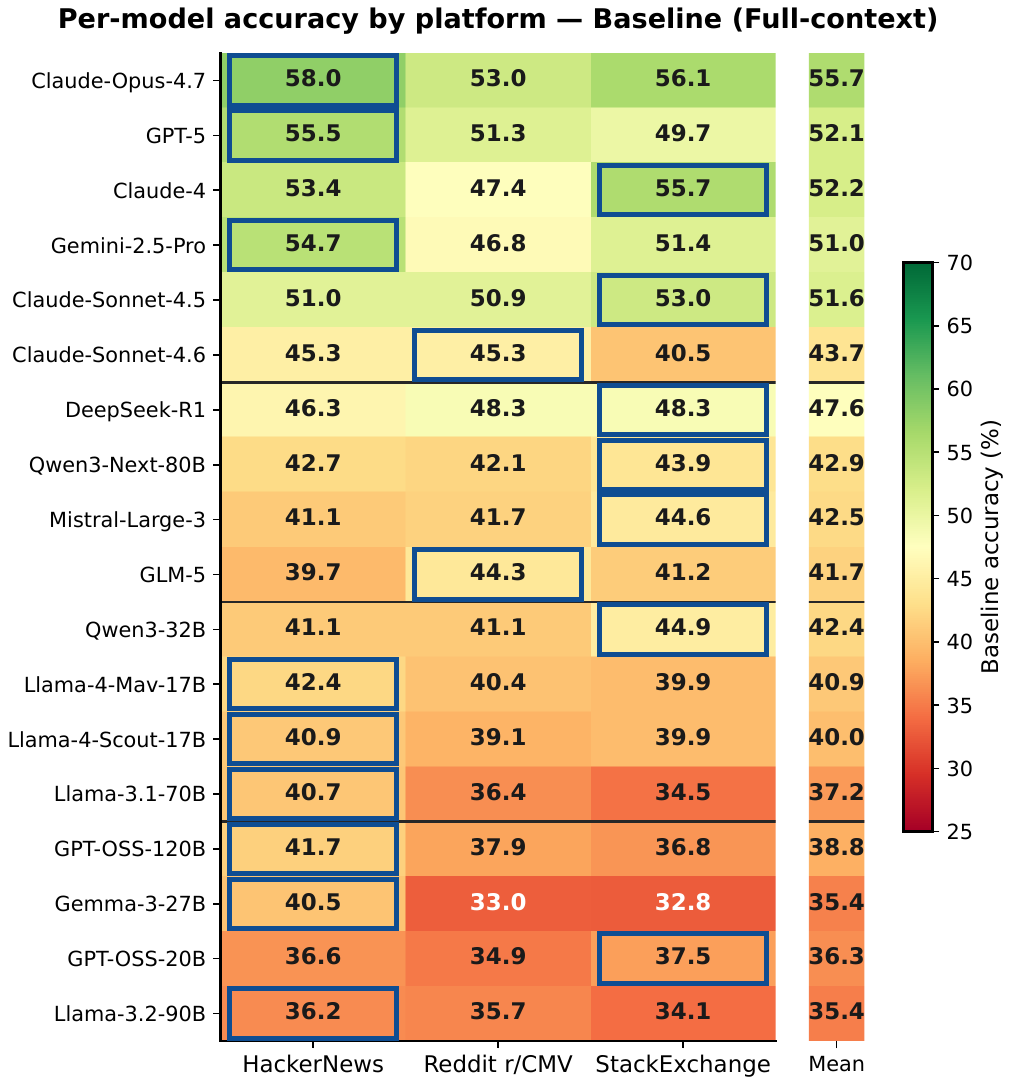}
  \caption{\textbf{Per-model accuracy by platform under Baseline.}
  Hacker News, Reddit r/CMV, Stack Exchange, and per-model mean are shown for each model.}
  \label{fig:domain-landscape-app}
\end{figure}

\begin{figure*}[!t]
  \centering
  \includegraphics[width=0.97\linewidth]{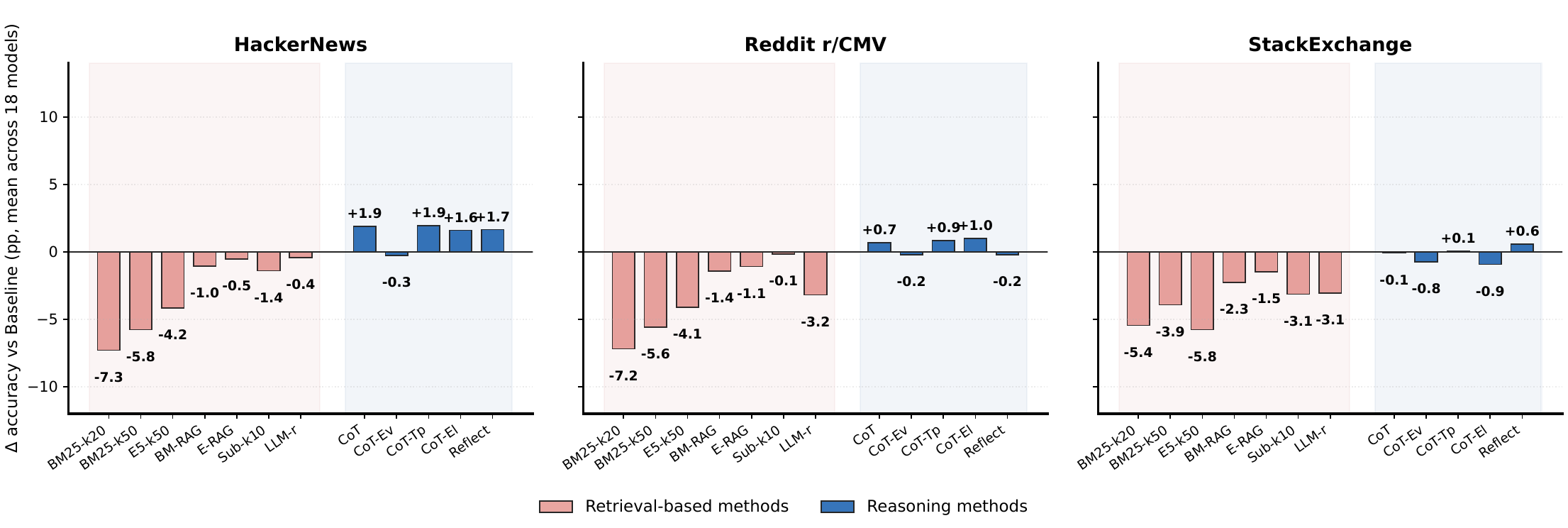}
  \caption{\textbf{Domain $\times$ method effect.}
  Each panel is one platform; bars are mean $\Delta$ accuracy vs.\ Baseline averaged across $18$ models.}
  \label{fig:domain-method-app}
\end{figure*}

Fig.~\ref{fig:capability-landscape-app} reports per-model capability accuracy under Baseline.
Figs.~\ref{fig:domain-landscape-app} and~\ref{fig:domain-method-app} provide the corresponding platform-level views.

\subsection{Reasoning Prompt Effects}
\label{app:reasoning-lift}

\begin{figure}[!t]
  \centering
  \includegraphics[width=0.99\linewidth]{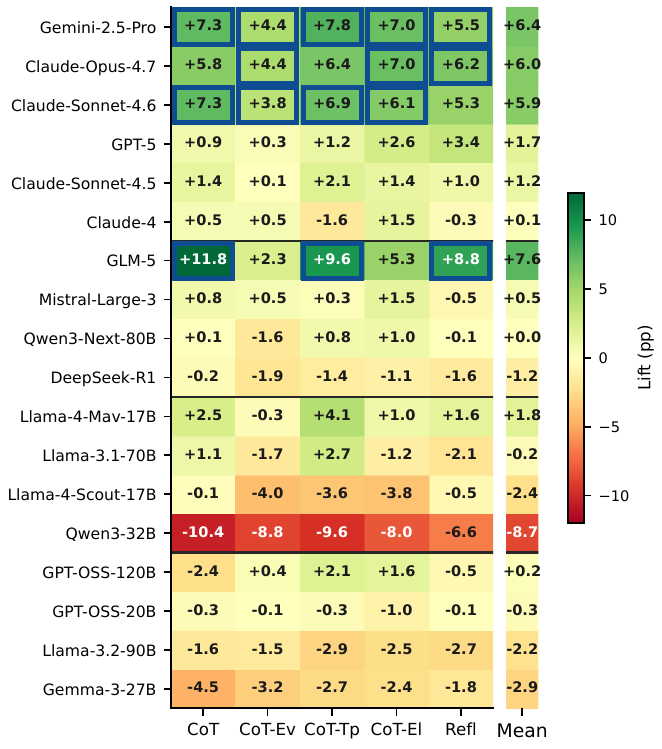}
  \caption{\textbf{Reasoning workflow lift over Baseline, per model.}
  Cells are accuracy under the named workflow minus Baseline in percentage points; the rightmost strip is the per-model mean lift.}
  \label{fig:reasoning-heatmap-app}
\end{figure}

Table~\ref{tab:reasoning-lift} summarizes per-model lift across reasoning-style workflows, while Fig.~\ref{fig:reasoning-heatmap-app} provides the corresponding visual comparison.

\begin{table*}[!t]
\small
\centering
\renewcommand{\arraystretch}{1.05}
\setlength{\tabcolsep}{6pt}
\caption{\textbf{Reasoning workflow lift over Baseline, per model.} Each cell is accuracy under the named workflow minus accuracy under Baseline (pp). Negative entries (\textcolor{red}{red}) indicate the workflow hurts the model. Models are sorted by mean lift descending. Reflect = self-reflection; CoT-Evid / CoT-Temp / CoT-Elim are chain-of-thought variants prompting evidence enumeration, temporal reasoning, or elimination of distractors, respectively.}
\label{tab:reasoning-lift}
\begin{tabular}{@{}l rrrrr r@{}}
\toprule
\textbf{Model}     & CoT                    & CoT-Evid              & CoT-Temp              & CoT-Elim              & Reflect               & \textbf{Mean} \\
\midrule
GLM-5              & +11.8                  & +2.3                  & +9.6                  & +5.3                  & +8.8                  & +7.6 \\
Gemini-2.5-Pro     & +7.3                   & +4.4                  & +7.8                  & +7.0                  & +5.5                  & +6.4 \\
Claude-Opus-4.7    & +5.8                   & +4.4                  & +6.4                  & +7.0                  & +6.2                  & +6.0 \\
Claude-Sonnet-4.6  & +7.3                   & +3.8                  & +6.9                  & +6.1                  & +5.3                  & +5.9 \\
Llama-4-Mav-17B    & +2.5                   & \textcolor{red}{-0.3} & +4.1                  & +1.0                  & +1.6                  & +1.8 \\
GPT-5              & +0.9                   & +0.3                  & +1.2                  & +2.6                  & +3.4                  & +1.7 \\
Claude-Sonnet-4.5  & +1.4                   & +0.1                  & +2.1                  & +1.4                  & +1.0                  & +1.2 \\
Mistral-Large-3    & +0.8                   & +0.5                  & +0.3                  & +1.5                  & \textcolor{red}{-0.5} & +0.5 \\
GPT-OSS-120B       & \textcolor{red}{-2.4}  & +0.4                  & +2.1                  & +1.6                  & \textcolor{red}{-0.5} & +0.2 \\
Claude-4           & +0.5                   & +0.5                  & \textcolor{red}{-1.6} & +1.5                  & \textcolor{red}{-0.3} & +0.1 \\
Qwen3-Next-80B     & +0.1                   & \textcolor{red}{-1.6} & +0.8                  & +1.0                  & \textcolor{red}{-0.1} & +0.0 \\
Llama-3.1-70B      & +1.1                   & \textcolor{red}{-1.7} & +2.7                  & \textcolor{red}{-1.2} & \textcolor{red}{-2.1} & \textcolor{red}{-0.2} \\
GPT-OSS-20B        & \textcolor{red}{-0.3}  & \textcolor{red}{-0.1} & \textcolor{red}{-0.3} & \textcolor{red}{-1.0} & \textcolor{red}{-0.1} & \textcolor{red}{-0.3} \\
DeepSeek-R1        & \textcolor{red}{-0.2}  & \textcolor{red}{-1.9} & \textcolor{red}{-1.4} & \textcolor{red}{-1.1} & \textcolor{red}{-1.6} & \textcolor{red}{-1.2} \\
Llama-3.2-90B      & \textcolor{red}{-1.6}  & \textcolor{red}{-1.5} & \textcolor{red}{-2.9} & \textcolor{red}{-2.5} & \textcolor{red}{-2.7} & \textcolor{red}{-2.2} \\
Llama-4-Scout-17B  & \textcolor{red}{-0.1}  & \textcolor{red}{-4.0} & \textcolor{red}{-3.6} & \textcolor{red}{-3.8} & \textcolor{red}{-0.5} & \textcolor{red}{-2.4} \\
Gemma-3-27B        & \textcolor{red}{-4.5}  & \textcolor{red}{-3.2} & \textcolor{red}{-2.7} & \textcolor{red}{-2.4} & \textcolor{red}{-1.8} & \textcolor{red}{-2.9} \\
Qwen3-32B          & \textcolor{red}{-10.4} & \textcolor{red}{-8.8} & \textcolor{red}{-9.6} & \textcolor{red}{-8.0} & \textcolor{red}{-6.6} & \textcolor{red}{-8.7} \\
\midrule
\rowcolor{accentcell}
\textbf{Mean (18)} & +1.1                   & \textcolor{red}{-0.3} & +1.2                  & +0.9                  & +0.8                  & +0.7 \\
\bottomrule
\end{tabular}
\end{table*}

\subsection{Retrieval Diagnostics}
\label{app:retrieval-diagnostics}

Table~\ref{tab:evidence-recall-app} measures whether standard comment-level retrievers recover the author-side evidence nodes used to justify gold answers.
For each item, recall is the fraction of cited evidence nodes that appear in the retrieved comment set; mean recall is averaged over the released items.

\begin{table}[!htbp]
\small
\centering
\renewcommand{\arraystretch}{1.05}
\setlength{\tabcolsep}{5pt}
\caption{\textbf{Evidence-node recall for standard comment-level retrievers.}
Values are mean recall against author-side evidence-node lists, averaged over the $1{,}462$ released items.
Only 8.8\% of items reach $\geq 75\%$ recall in the best of the four retrievers, and 3.5\% attain full coverage at the largest budget.}
\label{tab:evidence-recall-app}
\rowcolors{2}{rowshade}{white}
\begin{tabular}{@{}l c c@{}}
\toprule
\rowcolor{white}
\textbf{Retriever} & \textbf{Budget} & \textbf{Mean evidence recall} \\
\midrule
BM$25$             & $k{=}20$        & 0.201 \\
E$5$               & $k{=}20$        & 0.203 \\
BM$25$             & $k{=}50$        & 0.308 \\
E$5$               & $k{=}50$        & 0.310 \\
\bottomrule
\end{tabular}
\end{table}

\subsection{Validity and Robustness Checks}
\label{app:validity}

This subsection reports three complementary checks discussed in \S\ref{sec:results-validity}: shortcut robustness, contamination, and the relationship between episode length and accuracy.

\subsubsection{Shortcut Robustness}
\label{app:shortcut-robustness}

We run shortcut probes to check whether LongSocialBench can be solved from answer-option artifacts, question wording, topic familiarity, or bag-of-comments cues.
All probes are evaluated on stratified subsets by capability and source.
Table~\ref{tab:shortcut-robustness} summarizes the shortcut tested and the observed result.

\begin{table*}[!htbp]
\small
\centering
\renewcommand{\arraystretch}{1.10}
\setlength{\tabcolsep}{6pt}
\caption{\textbf{Shortcut robustness summary.}
Each probe removes or perturbs information that would be unnecessary if the corresponding shortcut were sufficient.
Accuracy values are means across $18$ models on the probe subsets; the full-benchmark Baseline is $43.9\%$ for context.
For open-answer, the table reports the same-item accuracy change and model-ranking correlation.}
\label{tab:shortcut-robustness}
\rowcolors{2}{rowshade}{white}
\begin{tabularx}{\linewidth}{@{}l l c X@{}}
\toprule
\rowcolor{white}
\textbf{Probe} & \textbf{Shortcut tested} & \textbf{Observed} & \textbf{Interpretation} \\
\midrule
Option-only
& option prior
& 31.5\%
& option text alone is insufficient \\

Question-only
& stem leakage
& 38.0\%
& question wording alone is insufficient \\

Title/root-only
& topic familiarity
& 35.4\%
& topic-level information is insufficient without replies \\

Shuffled thread
& bag-of-comments inference
& 40.4\%
& performance remains limited when local thread order is perturbed \\

Open-answer subset
& multiple-choice artifact
& $-7.6$ pp; $\rho{=}1.00$
& absolute accuracy drops, but the model ranking is preserved \\
\bottomrule
\end{tabularx}
\end{table*}

For the open-answer probe, each model receives the same thread, title, and question stem as the multiple-choice setting but \emph{without} the four answer options, and is asked to produce a short free-text answer.
The free-text response is content-scored against the gold-option text rather than parsed to a letter: a counting-style gold option such as ``C. 5 answers'' is credited if the response asserts the same count, and an interpretive gold option such as ``A. real-world example, formal framework, philosophical argument'' is credited if the response names the same items in the same ordering.
This evaluates whether the model can recover the correct content without selecting among A--D.

On the same items, open-answer accuracy is $7.6$ pp lower than multiple-choice, while the model ranking is preserved almost exactly (Spearman $\rho{=}1.000$ across the $18$ models).
Together, the reduced-input, shuffled-thread, and open-answer probes indicate that the benchmark is not solvable from answer options, question wording, topic cues, or bag-of-comments cues alone, and that the model ranking is not an artifact of the multiple-choice format.

\subsubsection{Pre- and Post-Cutoff Contamination Probe}
\label{app:contamination-split}

We test whether memorization of source pages inflates accuracy on platforms that overlap pretraining corpora, comparing two time-disjoint cohorts: Stack Exchange, collected from $2008$ to $2018$ archives, and Hacker News, collected from $2026$ Q1.
Table~\ref{tab:contamination-app} reports the per-tier Baseline means.
The pre-cutoff cohort shows no systematic advantage on any tier, providing no indication that source memorization explains the headline gap.
The probe cannot rule contamination out entirely.

\begin{table}[!htbp]
\small
\centering
\renewcommand{\arraystretch}{1.06}
\setlength{\tabcolsep}{4pt}
\caption{\textbf{Pre- and post-cutoff contamination probe.}
Mean Baseline accuracy by tier on the two time-disjoint cohorts; Reddit is reported for context.
$\Delta_{\text{HN}-\text{SE}}$ is the post-cutoff minus pre-cutoff gap.}
\label{tab:contamination-app}
\rowcolors{2}{rowshade}{white}
\begin{tabular}{@{}l c c c c@{}}
\toprule
\rowcolor{white}
\textbf{Tier}       & \textbf{HN}   & \textbf{Reddit} & \textbf{SE}   & \textbf{$\Delta_{\text{HN}-\text{SE}}$} \\
\midrule
Closed frontier (6) & 53.0          & 49.1            & 51.1          & $+1.9$ \\
Strong open (4)     & 42.4          & 44.1            & 44.5          & $-2.1$ \\
Mid open (4)        & 41.3          & 39.3            & 39.8          & $+1.5$ \\
Small (4)           & 38.8          & 35.4            & 35.3          & $+3.5$ \\
\midrule
\rowcolor{accentcell}
\textbf{All 18}     & \textbf{44.9} & \textbf{42.8}   & \textbf{43.6} & $\mathbf{+1.3}$ \\
\bottomrule
\end{tabular}\\[2pt]
{\footnotesize HN = Hacker News post-cutoff cohort, $2026$ Q1. SE = Stack Exchange
pre-cutoff cohort, $2008$ to $2018$. Reddit r/CMV spans both and is excluded from the
formal probe.}
\end{table}

\subsubsection{Episode-Length Effect}
\label{app:length-effect}

\begin{figure}[!htbp]
  \centering
  \includegraphics[width=\linewidth]{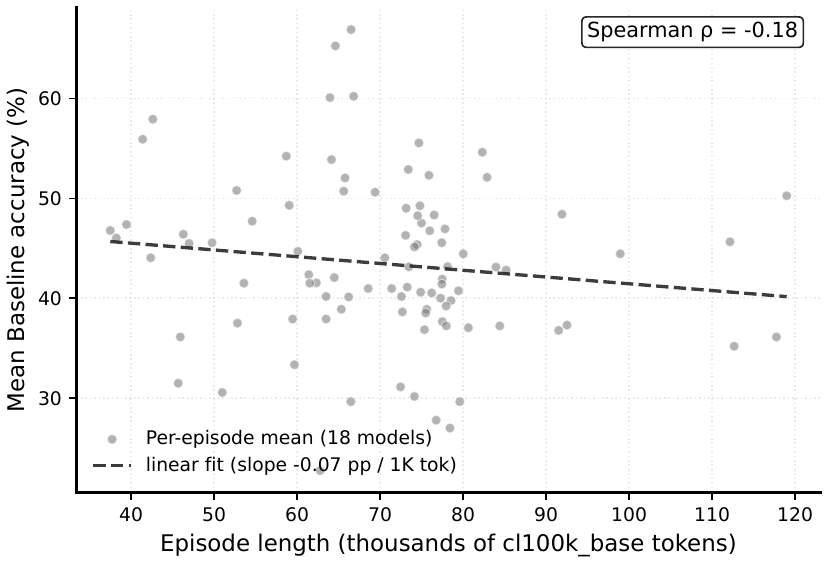}
  \caption{\textbf{Episode length vs.\ Baseline accuracy.}
  Each point is one of the $94$ released episodes: mean Baseline accuracy across the $18$
  models against episode length in \texttt{cl100k\_base} tokens, with the least-squares fit
  and the pooled Spearman $\rho$.}
  \label{fig:length-vs-accuracy-app}
\end{figure}

Episodes vary in size, so we check whether accuracy simply tracks input length.
Figure~\ref{fig:length-vs-accuracy-app} plots per-episode mean Baseline accuracy against episode length; the least-squares fit has slope $-0.07$ pp per $1$K tokens and the Spearman rank correlation is $\rho{=}{-}0.18$.
Length is therefore only weakly anti-correlated with accuracy, far too weak to account for the roughly $28$ pp gap to the human reference.

\subsection{Scope and Component Diagnostics}
\label{app:component-diagnostics}

Per-tier Gold-Scope lift is reported in the main text (Table~\ref{tab:goldscope-main}).
Table~\ref{tab:component-oracle-app} reports the component oracle case study on the stratified subset.

\begin{table}[!htbp]
\small
\centering
\renewcommand{\arraystretch}{1.06}
\setlength{\tabcolsep}{5pt}
\caption{\textbf{Component oracle case study.}
Claude-Opus-$4.7$ on the stratified $30$-episode / $510$-item subset.}
\label{tab:component-oracle-app}
\rowcolors{2}{rowshade}{white}
\resizebox{\columnwidth}{!}{%
\begin{tabular}{@{}l c@{}}
\toprule
\rowcolor{white}
\textbf{Condition}                            & \textbf{Accuracy} \\
\midrule
CoT baseline                                  & $63.1$ \\
Gold scope                                    & $66.7$ \\
Gold per-comment labels                       & $81.4$ \\
Gold scope + gold labels                      & $98.6$ \\
Structured enumerate-label-aggregate pipeline & $70.0$ \\
\bottomrule
\end{tabular}%
}
\end{table}

\subsubsection{Residual Errors under Gold-Scope}
\label{app:goldscope-errors}

We manually inspect $50$ Claude-Opus-$4.7$ errors under Gold-Scope and categorize the immediate source of each failure (Table~\ref{tab:goldscope-errors}).

\begin{table}[H]
\small
\centering
\renewcommand{\arraystretch}{1.06}
\setlength{\tabcolsep}{6pt}
\caption{\textbf{Manual audit of $50$ Claude-Opus-$4.7$ Gold-Scope errors.}
Each error is assigned the immediate source of the failure; percentages describe the audited sample.}
\label{tab:goldscope-errors}
\rowcolors{2}{rowshade}{white}
\begin{tabular}{@{}l c@{}}
\toprule
\rowcolor{white}
\textbf{Error type}    & \textbf{Share (\%)} \\
\midrule
\rowcolor{accentcell}
\textbf{Criterion use} & $\mathbf{60}$ \\
Thread structure       & $16$ \\
Counting rule          & $14$ \\
Surface cue            & $10$ \\
\bottomrule
\end{tabular}
\end{table}

Criterion-use errors dominate: the model has access to the relevant region but applies the question-specific inclusion or interpretation rule too broadly or too narrowly.
\section{Representative Questions}
\label{app:samples}

This appendix presents one representative LongSocialBench item for each of the six capability surfaces, providing a worked example for every surface in Table~\ref{tab:category-main}.
The examples are chosen for readability rather than difficulty and illustrate the social-reading decision each capability is designed to test.
Question stems and answer options follow the released benchmark items, with usernames replaced by structural references where they are not essential; rationales are shortened summaries of the released criteria.

\newcommand{\examplecard}[5]{%
\par\vspace{8pt}\noindent
{\setlength{\fboxsep}{5pt}\colorbox{groupshade}{\parbox{\dimexpr\linewidth-2\fboxsep\relax}{\textbf{#1}}}}\par\nopagebreak\vspace{3pt}
\noindent\textbf{Episode topic.}~{\small #2}\par\vspace{2pt}
\noindent\textbf{Question.}~\emph{``#3''}\par\vspace{2pt}
\noindent\textbf{Options.}~#4\par\vspace{2pt}
\noindent\textbf{Gold and rationale.}~{\small #5}\par
\noindent\textcolor{accentcell}{\rule{\linewidth}{0.6pt}}\par
\vspace{2pt}
}

\subsection{Example 1: Local social meaning}

\examplecard
{C1: Local social meaning}
{A Hacker News thread on IPv6's thirtieth birthday and its still-incomplete adoption. One top-level comment blames the confusion on display in the comment section itself---in a forum whose users are supposedly more technically savvy than most---and draws seven direct replies.}
{One top-level comment argues that the confusion on show in this very discussion is why IPv6 has not been adopted. Which of its direct replies accepts that observation but aims it back at the people making it?}
{
A.~A request for concrete examples of the confusion being alleged \\[2pt]
\textbf{B.~A joke about how the commenters rate their own expertise} \\[2pt]
C.~A link to adoption figures set against the closing prediction \\[2pt]
D.~A claim that the real obstacle was the absence of any incentive
}
{\textbf{B.} The winning reply quotes the ``yawning chasm of gaps in knowledge'' line and answers only that there is a Dunning--Kruger joke to be made here: it grants the diagnosis and turns it on the commenters, the author of the diagnosis included. The other options name real replies in the same set, which ask which misconceptions are meant, answer the closing prediction with an adoption figure, or reject the causal story outright. Every option shares the thread's vocabulary, so the item turns on reading a one-line joke as agreement plus turnabout against the comment it sits under.}

\subsection{Example 2: Reply-pair interaction}

\examplecard
{C2: Reply-pair interaction}
{A Hacker News thread on a post titled ``Google is dead. Where do we go now?'' One commenter reports that they and the other small software-business owners they know have all watched paid search become less profitable, and that most have given up on it. Four people reply directly.}
{A comment reports that the small software-business owners its author knows have all found paid search less and less profitable. Which of its direct replies answers with the same kind of evidence but the opposite conclusion?}
{
A.~The reply describing the commenter's own slow decline and shrinking budget \\[2pt]
B.~The reply asking what other approaches the commenter uses instead \\[2pt]
\textbf{C.~The reply describing many owners in other lines of business whose results have held up} \\[2pt]
D.~The reply describing twenty years of running such campaigns and rising complexity
}
{\textbf{C.} The parent argues from a circle of acquaintances who all run the same kind of business and all report the same decline; the gold reply answers in exactly that currency---many non-software owners whose results have held up---matching the form of the evidence while reversing the verdict. Two of the remaining replies corroborate the parent from personal experience and the third asks a question instead of offering evidence at all. Choosing correctly requires reading each reply against what its parent claimed and how it claimed it, not judging any reply on its own.}

\subsection{Example 3: Collective discussion structure}

\examplecard
{C3: Collective discussion structure}
{A Hacker News thread about a macOS redesign that made window resizing harder. Mid-discussion, one top-level comment proposes crowdfunding billboards in Cupertino to tell Apple how badly it missed; ten replies land underneath it.}
{A top-level comment proposes crowdfunding a Cupertino billboard criticizing Tahoe. How do its direct replies split among endorsing the idea, pushing back, and moving to a different topic?}
{
\textbf{A.~Five endorse; three push back; two move elsewhere} \\[2pt]
B.~Six endorse; two push back; two move elsewhere \\[2pt]
C.~Three endorse; five push back; two move elsewhere \\[2pt]
D.~Four endorse; four push back; two move elsewhere
}
{\textbf{A.} Five replies join in---pledging \$25, quoting real rates for an electronic billboard on Highway 101, and riffing on what it should say---while three deflate the idea, arguing the money belongs with open-source projects, that outside pressure will not turn the ship around, or that Cupertino may not even have billboards. Two more give a general opinion of the redesign without taking any position on the proposal. The item turns on reading a joke as endorsement and a rhetorical question as refusal, then holding the whole reply set in view at once.}

\subsection{Example 4: Temporal dynamics}

\examplecard
{C4: Temporal dynamics}
{A Hacker News thread on a television showrunner telling viewers to switch off their sets' ``garbage'' picture and sound processing. The episode's longest reply chain opens with a viewer who needs subtitles to follow anything and runs several rounds on whether built-in speakers are to blame.}
{Walk down this episode's longest reply chain. At which position does the conversation permanently move away from the opening topic of audio intelligibility? Count the starting comment as position one.}
{
A.~Position four \quad
B.~Position six \quad
C.~Position ten \quad
\textbf{D.~Position eight}
}
{\textbf{D.} The first seven comments stay on one subject---subtitles, dialogue mixing, and whether a thin panel can produce intelligible sound---each turn answering the last. The eighth is a two-word correction of which service a show streams on, the first comment in the chain that says nothing about audio, and every later turn follows it into brand names becoming generic words rather than back. Answering requires walking the chain in order and separating the true point of no return from the later turn at which the new topic merely becomes obvious.}

\subsection{Example 5: Participant profile}

\examplecard
{C5: Participant profile}
{A ChangeMyView discussion arguing that nuclear energy is good, where the leading replies press on cost, waste, and build times. The pro-nuclear participant profiled in the question appears in five separate parts of the thread.}
{Among this participant's comments in the episode, how many use a concrete number, statistic, or calculation to rebut another comment?}
{
A.~Six comments \quad
B.~Five comments \quad
\textbf{C.~Three comments} \quad
D.~Four comments
}
{\textbf{C.} This participant writes seven comments spread across five sub-discussions, and the recognizable move is answering objections with hard figures: Germany's spending on renewables against a grid still far dirtier than France's, fifty gigawatt-hours of storage covering roughly seven minutes of average US load, and no recorded injuries from spent fuel alongside the eight-day half-life of iodine-131. The remaining four comments contain no figures at all. Reaching the answer means collecting one participant's comments from across a long thread and judging what each one does, rather than reading any single exchange.}

\subsection{Example 6: Cross-subdiscussion variation}

\examplecard
{C6: Cross-subdiscussion variation}
{A Hacker News thread on a blog post arguing that macOS Tahoe's new icons are indefensible. The episode's second-most prolific commenter appears in eight separate top-level threads---panning the system icon set in one, cataloguing broken permissions in another, and opening a third one themselves.}
{The episode's second-most prolific user writes comments across eight distinct top-level threads. In which thread does that user adopt a notably different stance from their pattern elsewhere---offering a conciliatory defence of macOS against a Linux comparison rather than criticizing Apple's design choices?}
{
A.~The thread where the opening comment tells people to click the snow buttons \\[2pt]
\textbf{B.~The thread whose opening comment calls Apple's designer worship the root problem} \\[2pt]
C.~The thread about peak desktop UI contenders from the pre-2000 era \\[2pt]
D.~The thread discussing the article's hamburger menu and user agreement joke
}
{\textbf{B.} In seven of the eight threads this commenter is among the harshest Apple critics in the episode, dismissing the system icons as ``cheap, lame, boring, flat little hieroglyphics'' and reporting that the permissions panel has been broken for several releases. Only under the comment blaming Apple's worship of its own designers does that commenter change sides, replying that no Linux desktop environment comes close to the macOS experience despite its flaws and regressions. The baseline has to be assembled from several unconnected sub-discussions before the exception is visible; no single thread contains enough to decide it.}
\section{Construction Workflow}
\label{app:portal}

This appendix summarizes the contributor-facing construction workflow and the portal controls that enforce role separation.
Contributors cannot verify or adjudicate items they authored, and human-reference readers do not participate in item construction.

\subsection{Annotation and Verification Protocol}
\label{app:guidelines}

Table~\ref{tab:annotation-protocol} summarizes the four stages of item construction, verification, and adjudication while keeping model-facing inputs separate from internal construction records.

\begin{table*}[!t]
\small
\centering
\renewcommand{\arraystretch}{1.12}
\setlength{\tabcolsep}{5pt}
\caption{\textbf{LongSocialBench annotation and verification protocol.}
Authoring records define the social criterion and supporting evidence; verification records test whether an independent reader can recover the gold option under the same input condition given to models.}
\label{tab:annotation-protocol}
\rowcolors{2}{rowshade}{white}
\begin{tabularx}{\linewidth}{@{}p{2.6cm} X X@{}}
\toprule
\rowcolor{white}
\textbf{Stage} & \textbf{Contributor task} & \textbf{Quality control} \\
\midrule

Authoring
& Read the complete discussion, choose a clear scope, and write the question stem, four plausible options, gold option, plain-language criterion, and evidence nodes.
& The item must require reasoning over the specified scope rather than topic familiarity or one salient comment. Distractors reflect plausible wrong readings, such as wrong scope, nearby count, boundary error, or near-synonym social label. \\

Artifact control
& Remove answer-bearing wording from stems and options, including exact quotes, usernames, redundant structural hints, and distinctive phrases.
& These controls reduce locator leakage, option-paraphrase matching, keyword shortcuts, and open-web contamination. Capability and item-form tags are retained only as metadata. \\

Independent verification
& A different contributor re-answers the item from the same model-facing input used in evaluation: serialized thread, reply structure, question stem, and four answer options.
& The verifier does not see the author's gold option, criterion notes, evidence-node list, or rationale. Agreement checks answer stability and scope stability. \\

Adjudication
& Phase-2 mismatches and author-flagged borderline cases receive a third-pass review.
& The item is confirmed, clarified, rewritten and re-verified, or removed. A released item must have independent agreement or a stable adjudicated answer. \\

\bottomrule
\end{tabularx}
\end{table*}

\section{Ethics, Licensing, and Contributor Protocol}
\label{app:ethics}

This appendix documents the data-handling, contributor, and intended-use considerations behind LongSocialBench.

\paragraph{Source data and licensing.}
The released episodes are drawn from three publicly accessible reply-threaded sources: Hacker News, Reddit r/ChangeMyView, and Stack Exchange.
Copyright and other rights in the source discussion content remain with the respective rightsholders and are subject to the applicable platform terms and licenses.
Stack Exchange content is subject to the applicable CC BY-SA version associated with each contribution.
We redistribute only what is needed for benchmark evaluation: the serialized thread text with reply structure used by the items, with usernames replaced by per-episode pseudonyms and personally identifying spans removed where surfaced during verification.
The benchmark questions, answer options, and labels created by the authors are released under CC BY 4.0; the release does not grant additional rights to third-party source content.

\paragraph{Contributors and readers.}
Contributor recruitment and tasking follow a written protocol with role separation between authoring, independent verification, and adjudication, as summarized in Appendix~\ref{app:portal}.
Contributors completed a short calibration step before authoring or verifying released items.
Contributor and human-reader participation was compensated at rates consistent with institutional norms for annotation work of comparable difficulty, and the study was conducted under the institutional process applicable to public-text annotation at our home institution.

\subsection{Contributor Pool and Authoring Guidelines}
\label{app:contributor-protocol}

The item-construction pool consisted of 20 invited contributors with prior experience reading and analyzing long-form English discussions, including graduate students and researchers in NLP, social computing, computational social science, HCI, or related areas. All contributors were fluent English readers and were instructed to author items from the supplied discussion episode alone, without using external knowledge or platform-specific assumptions. Before contributing released items, contributors completed a calibration step covering the six capability surfaces, evidence-node selection, option construction, and common shortcut risks. Human-reference readers used for the final answerability comparison were kept separate from item authors, verifiers, and adjudicators.

Contributors received written guidelines requiring each item to specify: (i) the relevant discussion scope, such as a comment, direct-reply set, subtree, chain, participant trajectory, or cross-branch comparison; (ii) one of the six capability surfaces; (iii) a four-option question with exactly one supported answer; (iv) a plain-language decision criterion; and (v) supporting evidence nodes. The guidelines instructed authors to avoid answer-revealing locators, exact quotes, usernames, distinctive phrases, and distractors that could be eliminated by wording alone. They also required distractors to reflect plausible wrong readings, such as wrong scope, wrong criterion, boundary errors, lexical matches, or aggregation mistakes.

\paragraph{Intended use and dual-use considerations.}
LongSocialBench is intended for the evaluation of long-context language models on reply-structured discussion understanding.
The capability surfaces (e.g., reply role, turning point, participant trajectory) are scoped to behavior observable in the released thread; they are not designed for identity inference, off-platform attribution, or individual-level profiling of participants outside the released episodes.
The released benchmark artifacts are intended to support research and evaluation, and we discourage applications that target individual users in deployed settings.
\section{Release and Reproducibility}
\label{app:reproducibility}

We release LongSocialBench v$1.0$ at
\url{https://huggingface.co/datasets/xinyiliu-research/LongSocialBench}
to support reproducible evaluation.

\paragraph{Released artifacts.}
The release includes:
(i) the $1{,}462$ verified items with question stems, four answer options, gold options, and capability labels; and
(ii) the $94$ serialized episodes with the reply-structure metadata underlying the benchmark.
The two components are linked by episode identifiers and provide the model-facing discussion context and gold answers required for benchmark evaluation.

\paragraph{Evaluation reproducibility.}
All model evaluations were run under the workflow settings documented in Appendix~\ref{app:workflow-settings}, using publicly accessible model endpoints or open-weight checkpoints.
The released benchmark data support evaluation under these settings using the same discussion episodes, questions, answer options, and gold answers reported in the paper.

\end{document}